\documentclass{article}

\usepackage{arxiv}

\usepackage[utf8]{inputenc} 
\usepackage[T1]{fontenc}    
\usepackage{hyperref}       
\usepackage{url}            
\usepackage{booktabs}       
\usepackage{amsfonts}       
\usepackage{nicefrac}       
\usepackage{microtype}      
\usepackage{lipsum}		
\usepackage{graphicx}
\usepackage{natbib}
\usepackage{doi}
\usepackage{float}
\usepackage{titlesec}

\AtBeginDocument{
  \setlength{\headheight}{22.37994pt}
  \addtolength{\topmargin}{-8.37994pt}
}

\title{An approach based on class activation maps for investigating the effects of data augmentation on neural networks for image classification}

\author{ \href{https://orcid.org/0009-0005-3042-9929}{\includegraphics[scale=0.06]{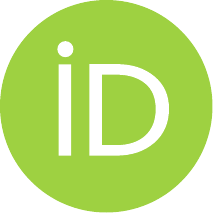}\hspace{1mm}Lucas M. Dorneles}\\
	Department of Computer Science\\
	Federal University of Rio Grande do Sul\\
	Porto Alegre, RS \\
	\texttt{luc.dorneles@gmail.com} \\
	\And
	\href{https://orcid.org/0000-0001-9328-9007}{\includegraphics[scale=0.06]{orcid.pdf}\hspace{1mm}Luan F. Garcia} \\
	Department of Computer Science\\
	Federal University of Rio Grande do Sul\\
	Porto Alegre, RS \\
	\texttt{luanfg@gmail.com} \\
    \And
	\href{https://orcid.org/0000-0002-4499-3601}{\includegraphics[scale=0.06]{orcid.pdf}\hspace{1mm}Joel L. Carbonera} \\
	Department of Computer Science\\
	Federal University of Rio Grande do Sul\\
	Porto Alegre, RS \\
	\texttt{joel.carbonera@inf.ufrgs.br} \\
}

\renewcommand{\headeright}{Technical Report}
\renewcommand{\undertitle}{Technical Report}
\renewcommand{\shorttitle}{An approach based on class activation maps for investigating the effects of data augmentation on neural networks for image classification}

\hypersetup{
pdftitle={An approach based on class activation maps for investigating the effects of data augmentation on neural networks for image classification},
pdfsubject={cs.LG, cs.CV},
pdfauthor={Lucas M.~Dorneles, Luan F.~Garcia, Joel L.~Carbonera},
pdfkeywords={Machine Learning, Explainability, Data Augmentation, Convolutional Neural Networks, Image Classification},
}

\begin{document}
\maketitle

\begin{abstract}
	Neural networks have become increasingly popular in the last few years as an effective tool for the task of image classification due to the impressive performance they have achieved on this task. In image classification tasks, it is common to use data augmentation strategies to increase the robustness of trained networks to changes in the input images and to avoid overfitting. Although data augmentation is a widely adopted technique, the literature lacks a body of research analyzing the effects data augmentation methods have on the patterns learned by neural network models working on complex datasets. The primary objective of this work is to propose a methodology and set of metrics that may allow a quantitative approach to analyzing the effects of data augmentation in convolutional networks applied to image classification. An important tool used in the proposed approach lies in the concept of class activation maps for said models, which allow us to identify and measure the importance these models assign to each individual pixel in an image when executing the classification task. From these maps, we may then extract metrics over the similarities and differences between maps generated by these models trained on a given dataset with different data augmentation strategies. Experiments made using this methodology suggest that the effects of these data augmentation techniques not only can be analyzed in this way but also allow us to identify different impact profiles over the trained models.
\end{abstract}


\section{Introduction}
Neural networks have demonstrated surprising performance in various data processing tasks over the last decade, especially in the field of image classification \citep{todescato2023multiscale,todescato2024multiscale,pulsICEIS2024}. Their high performance and capacity to learn complex patterns from training data are impressive, and both their customizability and ease of use help make them a highly active field of research.

\clearpage

When engaging with image classification problems, it is common to use data augmentation techniques to increase the robustness of the model to changes in the input images and also to avoid overfitting \citep{won2023analyzing, shorten2019survey}. This increased robustness often means an increase in performance, making a model effective against a larger pool of input data. However, even with data augmentation being a widely adopted technique, there still lacks a body of research investigating the effects of data augmentation methods on the patterns learned by neural network models, especially those models working on more complex datasets. As such, understanding how data augmentation may alter the classification process of a neural network model may help us to select suitable data augmentation strategies for a given use case, or to provide insights for creating more robust and effective data augmentation techniques.

Analyzing the effects of data augmentation techniques, however, is not an easy nor straightforward task. Due to the nature of neural networks themselves, the process by which models learn to identify patterns in images, and how they use these patterns to classify the image, works in a manner that is hard to interpret. It is possible to analyze the weights inside the model (which may number in the millions), but they do not directly provide methods to comprehend the inner workings of neural networks and generate human-interpretable explanations for their choices. This makes analyzing neural networks a nontrivial task and a significant challenge in the field \citep{angelov2020towards}, lending it a reputation for working in a manner akin to a black box. 


In recent years, this black box property of neural networks has steadily given rise to the field of neural network interpretability, or explainability \citep{zhang2021survey, liu2018interpretable}. One of the goals in this field is to understand why neural network models provide a given output in a given context and strive to make a model's choices explainable to humans. These challenges also directly relate to the trustworthiness of neural network models in society; as their adoption in various fields increases, so does the burden of guaranteeing accountability for the choices they make \citep{liu2018interpretable, haar2023analysis, zhang2021survey}. Work in the field is ongoing, however, and there are still no out-of-the-box, easy-to-use methods to interpret neural networks and their choice patterns \citep{fan2021interpretability, muhammad2020eigen, olah2018building}.

Similarly, understanding how data augmentation itself may affect neural networks is still a complex issue \citep{tang2020explaining, won2023analyzing}. Most works in this field are interested in discerning the impact of data augmentation in the performance of the models, comparing performance metrics such as accuracy and recall to measure their effectiveness concerning classification tasks. However, due mainly to the difficulties in interpreting neural networks, it is not trivial to determine how different techniques of data augmentation may affect the way neural networks work \citep{arthur2022impact}, and especially to do so at scale.


When analyzing data augmentation techniques, it is common to evaluate their impacts in one of two ways: either by considering performance metrics of models trained on differently augmented datasets in a given classification task \citep{o2019comparing, nanni2021comparison, perez2017effectiveness, li2018data, chen2020gridmask}, or by qualitative image analysis by applying semi-supervised analysis techniques over artifacts generated by explainability techniques \citep{cao2022survey, won2023analyzing, uddin2020saliencymix}.

The first type of analysis compares the performance metrics (accuracy, F1-score, etc.) of models trained on differently augmented datasets. This analysis type investigates the impact that data augmentation techniques have quantitatively and at scale, but can only measure impact as understood by classification performance. This fact means it cannot offer insights into how the application of data augmentation impacted how the model processed each image during classification.

The second type of analysis works through the visual identification of important image regions. Through explainability techniques, such as \emph{class activation maps} (CAM) \citep{selvaraju2017grad}, we may generate artifacts that can help us understand how an image is being used in a classification task by a neural network. This method by itself is usually adopted in a qualitative approach where a human must visually compare the artifacts for models with and without augmentation to determine the most important regions for both models, given an input image. An alternative technique involves manually annotating input images to provide some ground truth for region importance. A significant drawback to this type of technique, however, is its difficulty in scaling up. Analyzing hundreds or thousands of images would take an unreasonable amount of human effort in visually analyzing CAM pairs or annotating input data.

Considering the limitations present in both common types of approaches, we identify a lack of scalable methods to obtain a more general and quantitative perspective of how different strategies of data augmentation impact the patterns a neural network learns while processing images, especially one that may apply to large image datasets without the need for human supervision.

In this work, we propose a methodology that combines the advantages of both of these common analysis types. The proposed approach uses explainability techniques, such as CAMs, to compare how the application of (or lack thereof) augmentation techniques affects the attribution of importance by a model to different image regions, doing so at scale and without the need for extensive human intervention. We hypothesize that by analyzing CAMs not in isolation but instead comparing maps generated by a baseline model against those generated by differently augmented models, we may be able to develop and use similarity metrics that help measure these differences in a scalable way. This approach should allow for a quantitative, scalable way to analyze the impacts of data augmentation on model behavior by comparing CAMs.


The results we obtained using the methodology proposed in the following sections are promising. Although there remains difficulty in obtaining clear-cut answers using the metrics we chose, we can garner some insights about the models trained with augmented data. Additionally, investigating the correlations between the metrics of different augmentations also suggests the existence of separate profiles of impact across the data augmentation techniques we chose. In essence, this work attempts to tackle a single question: how can one use CAMs to automatically and quantitatively analyze the effects of data augmentation on a neural network model?

This work furthers the research into that question by proposing a basic, generically structured methodology to quantitatively evaluate the impacts of data augmentation in convolutional neural networks dealing with image classification tasks. Additionally, this work contributes by proposing an initial set of metrics that may be used within that methodology, as well as demonstrating the application of the methodology for a specific selection of metrics and augmentations, using the Grad-CAM \citep{selvaraju2017grad} method of generating CAMs, applied to the CIFAR-10 dataset \citep{krizhevsky2009learning} and adopting the EfficientNet B0 \citep{tan2019efficientnet} neural network architecture. Another quality of this methodology is its extensibility; future works may also expand on this methodology, including but not limited to using more complex metrics and proposing new ways to analyze said metrics, or modify aspects of this methodology to fit other image-classification-related needs. 

 



This work is structured as follows. In \hyperref[ch:bg]{Section \ref*{ch:bg}}, we will expand on fundamental concepts to the understanding of the proposed methodology. In \hyperref[ch:related]{Section \ref*{ch:related}}, we discuss works related to our goals. In \hyperref[ch:methodology]{Section \ref*{ch:methodology}}, we present the proposed methodology, covering its' steps and their explanations. In \hyperref[ch:experiments]{Section \ref*{ch:experiments}}, we discuss how the proposed methodology was applied in a specific context, as well as showcasing and analyzing our obtained results. 

\section{Background}\label{ch:bg}

The main concepts needed to understand the methodology proposed in this work are covered in the following subsections. First, we will introduce the concepts of machine learning and neural networks (\hyperref[ch:bg:ml]{Section \ref*{ch:bg:ml}}). Then, we will cover the topics of data augmentation (\hyperref[ch:bg:da]{Section \ref*{ch:bg:da}}) and model performance metrics (\hyperref[ch:bg:pm]{Section \ref*{ch:bg:pm}}). Finally, we will explain the concept of CAMs (\hyperref[ch:bg:gradcam]{Section \ref*{ch:bg:gradcam}}), a central concept in our work. 


\subsection{Machine Learning}\label{ch:bg:ml}

Machine Learning (ML) is a branch of Artificial Intelligence (AI) that focuses on developing algorithms capable of improving their performance through experience and learning from data \citep{mitchell1997introduction}. 
Traditionally, ML is divided into three broad categories: Supervised learning, unsupervised learning, and reinforcement learning \citep{murphy2012machine}. This work focuses on supervised learning. In supervised learning, there are two main types of tasks: regression and classification. Regression tasks involve training a model to learn a mapping function between input data and continuous output values based on labeled input-output pairs \citep{patterson2017deep}. Classification tasks, which are the focus of this work, have the objective of finding a mapping function between input data and discrete target labels \citep{patterson2017deep}.

For our work, the input data we want to classify are images, and the labels are the different types of entities (objects, animals, etc) represented in the images. To do this, we make use of artificial neural networks.




Artificial Neural Networks, or simply Neural Networks, constitute a specific approach to ML. They consist of algorithms inspired by the structure of human brains, following a hypothesis that the layered, neuron-based structure of the human brain may be an effective structure for mimicking intelligent behavior \citep{murphy2012machine}. Neural networks are part of a broad family of techniques for machine learning called Deep Learning, in which hypotheses take the form of complex algebraic circuits with tunable connection strengths \citep{DBLP:books/aw/RN2020}. In this sense, artificial neural networks are structured as interconnected layers of simple processing units (which represent neurons) that have weights associated with the connections between units of adjacent layers. Each node in a layer is capable of processing inputs and producing outputs which may then be sent to nodes further in the chain of layers, similar to how, in the human brain, a signal may travel through a chain of neurons. In practice, however, the actual resemblance between neural networks and neural cells and structures is only superficial \citep{DBLP:books/aw/RN2020}.

Training artificial neural network models commonly involves applying three main steps: forward propagation, back-propagation, and gradient descent. Forward propagation consists of the process of passing the information through the layers of nodes, processing the initial input (images in our case) until it arrives at the final layer  \citep{goodfellow2016deep}. Each layer of the neural network applies linear and non-linear transformations to the data it receives, propagating the processed signal forward until the final output is produced. From this final output, we determine the error produced between the expected value and the value obtained by the model. Determining the error is done by a cost function (also known as loss function) that measures this difference in some way, and many cost functions may be used for this purpose. 

Back-propagation refers to the algorithm that is executed after the final layer output is generated. It propagates backward the cost function gradients in relation to the model parameters (the weights between nodes and biases associated with each node). These gradients may be understood as the contribution of each parameter to the final error produced by the model, or the sensibility of the network to that parameter; the higher, the more it is considered to have contributed to the error, and the more its value needs to be adjusted \citep{goodfellow2016deep}.

Gradient descent is then applied after the parameter gradients have been set. It consists of an optimization algorithm used to adjust the parameters of the model in the opposite direction to the gradients calculated during back-propagation (the gradients may also be understood as first-order derivatives), intending to minimize the cost function \citep{goodfellow2016deep}. 



Specifically, in this work, we use a kind of artificial neural network called Convolutional Neural Networks (CNN). A CNN is a network that contains spatially local connections, at least in the early layers, and has patterns of weights that are replicated across the units in each layer \citep{DBLP:books/aw/RN2020}. A pattern of weights that is replicated across multiple local regions is called a kernel and the process of applying the kernel to the pixels of the image (or to spatially organized units in a subsequent layer) is called convolution \citep{DBLP:books/aw/RN2020}. 




\subsection{Data Augmentation}\label{ch:bg:da}

Data augmentation is one of the fundamental techniques this work is based on. They refer to a subset of regularization techniques, which work by introducing additional information to the ML model to better capture some properties of the problem being modeled \citep{khalifa2022comprehensive}. Image data augmentation, in particular, involves introducing variability in the training dataset itself, generating new data based on the original training images. This is done to increase the robustness of the model to variations in the input data.

In the realm of data augmentation, several different techniques may be used to generate augmented images from training images. As it would be prohibitively time-consuming to test all of the most popular data augmentation techniques for this work, we decided to focus on seven techniques that represent a wide variety of different data augmentation approaches. These are:

\begin{itemize}
  \item Affine transformation, which involves applying various linear and non-linear transformations (rotation, translation, scaling, shearing) in conjunction to one image. An example can be seen in \autoref{fig:aug:affine}.

    \begin{figure}[H]
    \caption{Example of affine transformation augmentation.}
    \label{fig:aug:affine}
    \includegraphics[height=4cm]{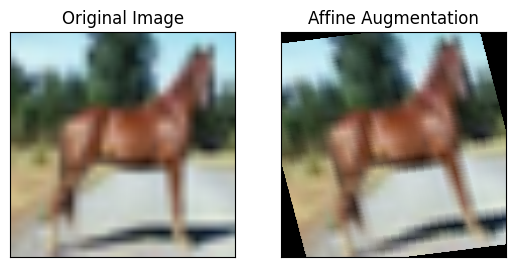}
    \centering
    \end{figure}
  
  \item Cutmix, which involves cutting a section of an image B with class Y and mixing it with an image A, which has a class X. The resulting image is then considered as having class X. An example is shown in \autoref{fig:aug:cutmix}.

    \begin{figure}[H]
    \caption{Example of cutmix augmentation.}
    \label{fig:aug:cutmix}
    \includegraphics[height=4cm]{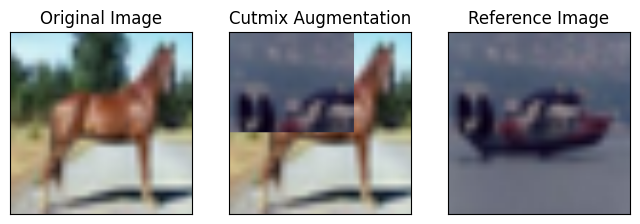}
    \centering
    \end{figure}
  
  \item Color Jitter, which randomly changes the brightness, contrast, saturation, and hue of an image. An example is shown in \autoref{fig:aug:color}

    \begin{figure}[H]
    \caption{Example of color jitter augmentation.}
    \label{fig:aug:color}
    \includegraphics[height=4cm]{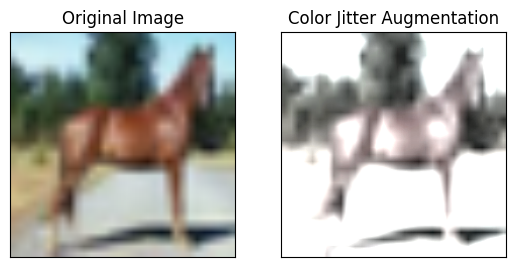}
    \centering
    \end{figure}

  \item Elastic transforms, a transform that works based on displacement vectors applied to all pixels. An example is shown in \autoref{fig:aug:elastic}.

    \begin{figure}[H]
    \caption{Example of elastic transformation augmentation.}
    \label{fig:aug:elastic}
    \includegraphics[height=4cm]{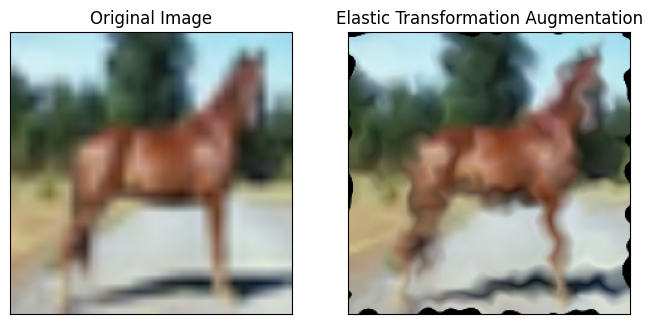}
    \centering
    \end{figure}
  
  \item Equalization, in which the histograms of each color channel of an image are equalized. Example in \autoref{fig:aug:eq}

    \begin{figure}[H]
    \caption{Example of equalization augmentation.}
    \label{fig:aug:eq}
    \includegraphics[height=4cm]{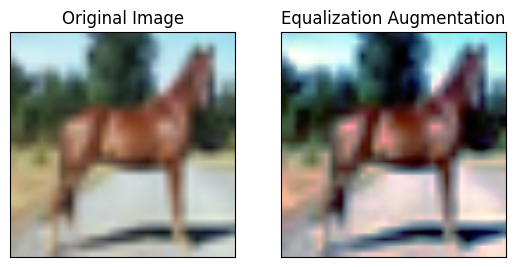}
    \centering
    \end{figure}

  \item Gaussian Blur, which applies Gaussian blurring to an image. An example is shown in \autoref{fig:aug:gaussblur}

    \begin{figure}[H]
    \caption{Example of Gaussian blur augmentation.}
    \label{fig:aug:gaussblur}
    \includegraphics[height=4cm]{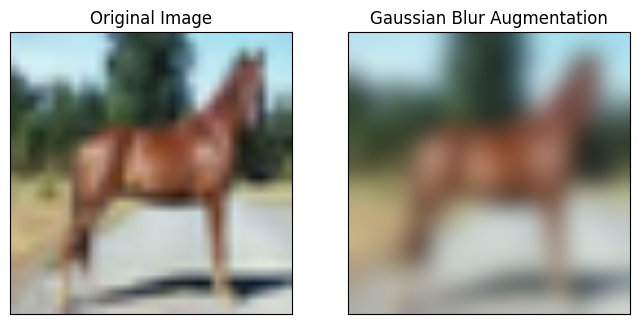}
    \centering
    \end{figure}

  \item Random Cropping, which crops the input image at a random location and resizes the cropped area to be the same size as the input image's. An example is shown in \autoref{fig:aug:randcrop}

    \begin{figure}[H]
    \caption{Example of random cropping augmentation.}
    \label{fig:aug:randcrop}
    \includegraphics[height=4cm]{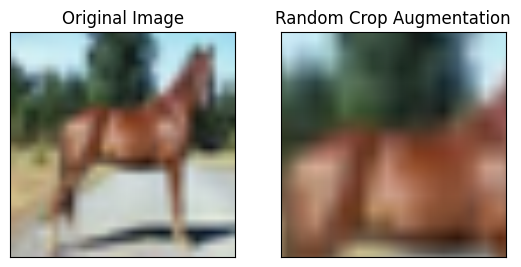}
    \centering
    \end{figure}
  
\end{itemize}

\subsection{Performance Metrics}\label{ch:bg:pm}
To evaluate the performance of CNNs, it is common to track metrics related to the correct and incorrect predictions made by the model. To track the performance of the models used in this work, we use accuracy, precision, F1-score, and recall. As we are dealing with a multi-class problem, we use the macro averages of these metrics.

Accuracy, precision, F1-score, and recall are calculated based on the confusion matrix resulting from model inferences. Confusion matrices refer to the relationship between ground truth labels and predicted labels in classification problems. To trace these relationships, we rely on the concepts of true positives (TPs), true negatives (TNs), false positives (FPs), and false negatives (FNs). In classification tasks with only two target classes, positive represents the presence of a characteristic, and negative represents the lack of said characteristic. TPs refer to instances where the model predicts positively and the correct label is also positive; TNs refer to cases where the prediction was negative and the correct label was also negative; FPs refer to the model predicting positively but the correct label being negative; and FNs referring to cases where the model predicted negatively but the correct label is positive. By counting the number of instances of each of these cases, we may create a confusion matrix \citep{bishop2006pattern}. In multi-class problems with $N$ classes, such as the one we are tackling, this matrix has an $N$ by $N$ shape, where the $N$ lines and columns represent each of the $N$ classes. The macro-average for a given metric is taken by averaging the sum of that metric over all $N$ classes.

\begin{itemize}
    \item Accuracy measures how often a classification ML model makes correct predictions overall.

\begin{center}
    $Accuracy = \frac{TP+TN}{TP+TN+FP+FN}$
\end{center}
    \item Precision measures the ratio of correct positive predictions made against all positive predictions made by the model.

\begin{center}
    $Precision = \frac{TP}{TP+FP}$
\end{center}
\item Recall measures the ratio of correct positive predictions made against all instances that possess the characteristic.

\begin{center}
    $Recall = \frac{TP}{TP+FN}$
\end{center}
\item F1-Score combines precision and recall into a single metric, providing a way to assess the impact of both false positives and false negatives simultaneously.

\begin{center}
    $F1-Score = \frac{2 \times (Recall \times Precision)}{Recall+Precision}$
\end{center}
\end{itemize}

\subsection{Grad-CAM and Class Activation Maps}\label{ch:bg:gradcam}
Explainability is still a significant challenge for neural networks. When models are trained, the patterns they learn are represented by the internal weights of the model, which may number in the millions or even billions. This makes analyzing image classification outputs of a model prohibitively complex if one were to inspect only a model's weights.

As tools to alleviate this problem, many techniques have been developed over the years. Deconvolutional neural networks, for example, can provide dense activation maps of an image by applying deconvolution operations and unpooling layers to the features extracted by CNNs \citep{haar2023analysis}; layer-wise relevance propagation (LWRP) can provide activation maps that not only indicate which parts of an image were important for classification but also which pixels support a different classification \citep{haar2023analysis}; and inversions allow recreation of the most important parts of an image by inverting what each layer of a model filters out to see what it considered important at that layer \citep{haar2023analysis}.

Amongst these, one of the most commonly used techniques to explain neural network classifications is that of CAMs. CAMs are a type of explainability method, commonly considered a specific type of saliency map, that indicates the discriminative image regions used by the neural network to identify that category \citep{zhou2016learning}. This explainability technique attributes the importance or contribution of each pixel of an input image to the final classification generated by the neural network. CAMs help establish how neural network models attribute importance to the different regions of an image \citep{zhou2016learning}, allowing visual analysis of what models consider important image regions.

\begin{figure}[H]
\caption{Example of CAM generated by Grad-CAM using the baseline model we trained as the selected model, the last convolutional layer as the target layer, and the predicted class as the target class.}
\label{fig:cam}
\includegraphics[height=4cm]{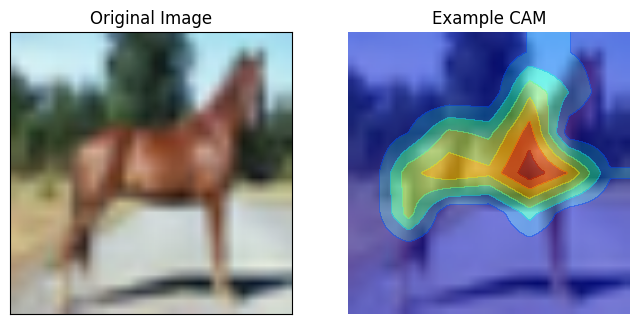}
\centering
\end{figure}

There are many ways of generating CAMs, such as Score-CAM \citep{wang2020score}, LayerCAM \citep{jiang2021layercam}, and Eigen-CAM \citep{muhammad2020eigen}, but the method we will be using in this work is the Gradient-Weighted Class Activation Map (Grad-CAM) \citep{selvaraju2017grad}. Grad-CAM uses the gradients of any target label (for example, ‘dog’ in a classification network) flowing into the last convolutional layer of a model to produce a CAM highlighting the important regions in the image for predicting the label. Given an image, a target layer, usually (but not necessarily) the final convolutional layer, and a target class, it forward-propagates the input through the CNN part of the model and obtains the raw score for the target category. The gradients of all outputs are then set to zero except for the desired class, which is set to 1. The result obtained by forward-propagation is then back-propagated up to the target layer that has the convolutional feature maps of interest, which are then combined to compute the coarse Grad-CAM localization. This localization represents which areas the model most strongly influences it to make a particular decision, meaning that a disturbance in that area would more strongly affect the model outcome for that class. Another interpretation of CAMs is that they represent the areas of the image that the model is the most sensitive to for a given class. Lastly, as the visualization generated is coarse, it is upscaled to the image size by pointwise multiplication with guided back-propagation to generate the final class activation map. The resulting mappings will have pixels normalized in the $[0, 1]$ range. An example of this type of CAM can be seen in \autoref{fig:cam}. These maps will be one of the main tools utilized in this work.

\section{Related Works}\label{ch:related}

Although the fields of neural network explainability and data augmentation have become increasingly popular over time, the literature still lacks a body of work analyzing how data augmentation may affect explainability. Among those works that exist, here are a few that explore a field of research similar to ours.

Firstly, there is \citep{tang2020explaining}, which inspired our work and whose investigations on data augmentation and explainability we follow up on. Their work explores the impact of five different data augmentation methods on the CAMs of the models trained with augmented data in an image classification task. Their process is similar to ours, conducting an initial foray in constructing quantitative methods for evaluating data augmentation impacts, but doing so in a more specific way. The work made by \citep{tang2020explaining} has its limitations, however. The authors selected MNIST, a simplistic grayscale dataset as the basis of their analysis. They also only trained one model per augmentation, subjecting the results to higher statistical randomness related to the impacts of starting states for the augmentation methods and neural networks. Our work improves on their initial exploration by using a more varied set of similarity metrics between CAMs; adopting CIFAR10 \citep{krizhevsky2009learning}, a significantly more complex colored dataset, as the basis for the experiments; and training multiple models per augmentation to ensure higher statistical robustness of the results. Our work also expands on theirs by using more augmentation methods and measuring other similarity metrics between CAMs, as well as analyzing correlations between the metrics taken as they co-vary over the test dataset. Our work also differs from theirs by providing a structured methodology that may be applied to a large variety of image datasets, methods of generating CAMs, sets of augmentations, sets of similarity metrics between CAMs, and CNN architectures, the choice of architectures being limited by compatibility with CAMs.

In \citep{won2023analyzing}, the authors also investigate the impacts of data augmentation on the interpretability of the models generated, but using a different methodology. It uses Grad-CAM and Information Bottlenecks through Attribution  \citep{schulz2020restricting} to generate attribution maps for images, which are then used to compare the CAMs generated by the different augmentation qualitatively to one another. Though it is thematically very close to our work, their work is interested in measuring how data augmentation affects interpretability as understood by qualitative measures (Human-model alignment, Faithfulness, and Human-understandable concepts,) whereas our work is interested in the use of quantitative measures that may be applied automatically over large datasets in order to compare the impacts of different augmentations.

Other related works cover only tangentially the matter of analyzing image data augmentation impact and explainability. In \citep{uddin2020saliencymix}, for example, the authors investigate differences between the CAMs of different augmentations, analyzing their performance while introducing their own augmentation method. However, it is difficult to find works where the main focus is specifically on explainability with data augmentation, as a majority of articles that analyze and compare data augmentation methods are more interested in comparing their performances, such as \citep{chen2020gridmask,radhakrishnan2017patchnet,yang2022image}.

\section{Proposed Methodology}\label{ch:methodology}

In this section, we explain our proposed methodology for generating quantitative analyses from CAMs in image classification tasks. For clarity and brevity purposes, from here on out, we will be referring to the models trained on augmented datasets as 'augmented models', and the model trained on the original CIFAR10 dataset (without applying any kind of augmentation) as 'baseline model'. Our main objective with this methodology is to propose a series of steps to analyze the impact of data augmentation on models trained on augmented data. To clarify, we are not interested in analyzing these augmented models or their artifacts in isolation, but rather by evaluating these models relative to the baseline model. Our intended analysis consists of comparing how data augmentation affects models in comparison to a baseline. In this way, our focus is on evaluating how the augmented models diverge from and co-vary with the baseline model. This approach allows us to collect metrics related to how the augmented models' CAMs diverge from the ones generated by the baseline model, and thus provides a quantitative analysis of their divergences.

The methodology consists of the following steps:
\begin{enumerate}
    \item Select a set of data augmentation techniques $A$ to be analyzed, and a set of starting states $X$. A starting state may be understood as some selection of variables or parameters that results in a specific initial configuration for the data augmentation methods, such as the seed given to a random number generator in the case of a data augmentation method that uses randomness. Note that the actual parameters of the augmentation itself (such as size of the crop for Random Cropping, standard deviation and kernel size for Gaussian blur, or the alpha and sigma for the Elastic Transformation method) should be the same across the $|X|$ states.
    \item Select a CNN architecture and base dataset.
    \item Select a method for generating CAMs.
    \item Select a set of metric functions $ME$ to be applied to the baseline and augmented CAMs.
    \item Select a set of methods to be applied to the metrics for analysis.
    \item Separate the dataset into training, test, and validation sets. The training, test, and validation sets must be stable across all models. For clarity in later definitions, define the test set as $I$.
    \item Train baseline model $B$ of the selected CNN architecture based on the selected training and validation sets. The validation sets are used to monitor the model training and extract the version of the model with the best performance against the validation set across all epochs, and they are used in this way for every single model. 
    \item Train $|X|$ versions of the chosen architecture for each augmentation, each of the $|X|$ models for a given augmentation $a$ with starting state $x\in X$, and training on their respective augmented dataset. This will result in $|X|\times|A|$ augmented models $M_{a, x}, a\in A, \;x\in X$. All models, including the baseline, should be instantiated with the same starting weights. This step is done to eliminate other sources of impact from the model results, such that the differences between the models are borne uniquely from the application of different augmentations. The set of augmented models will be defined as $M_{Aug} = M_{a,x}, \forall a\in A, \;\forall x\in X$, and the set of all models as $M = M_{Aug} \cup B$.
    \item Collect performance metrics (precision, recall, etc) for each of the models.
    \item Generate CAMs according to the selected method, generating augmented model CAM sets $S_{m}, \forall m\in M$. Defining the CAM-generating function as $f$, which takes a model $m$ and test image $i\in I$, a single CAM will be defined as $s_{m, i} = f(m,i), m\in M, \;i\in I$. The target layer for all models will be the last convolutional layer of the chosen architecture, and the target class will be the class predicted by the model for the respective test image. The CAM set $S_{m}$ for model $m$ will be defined as $S_{m} = s_{m, i}, m\in M, \;\forall i\in I$, meaning every individual CAM will be based off a different image in the test set, covering all of the test set. 
    \item Generate similarity metrics $model\_metrics\_set_{me,m}, \forall me\in ME, \;\forall m\in M_{Aug}$ between the augmented CAM sets $S_{m}, m\in M_{Aug}$ and baseline CAM set $S_{B}$. To do this, define the metric function $me\in ME$ between two individual CAMs as $me(s_{m,i}, s_{B, i}), m\in M_{Aug}, \;i\in I$. The set of metrics $model\_metrics\_set_{me, m}$ between the CAMs of a given augmented model $m$ and those of baseline model $B$ for a given metric type $me$ will be defined as $model\_metrics\_set_{me,m} =me(s_{m,i}, s_{B, i}), me\in ME, \;m\in M_{Aug}, \;\forall i\in I$.
    \item Average the metrics for each augmentation $a\in A$. The previous step generates $|X|$ values for a metric $me$, for each augmentation $a$, for a specific image $i\in I$. In this step, we aggregate these values in a mean, such that the value result for an aggregate metric function $me_{aggr}(a,i)$ based on metric function $me$ for a given augmentation method $a$ and test image $i$ is defined by $me_{aggr}(a, i) = \sum_{x\in X} me(s_{M_{a, x},i}, s_{B, i}) \div |X|, a\in A, \;i\in I$. The set of aggregate metric values for an augmentation method $a$ and metric type $me$ is defined as $aug\_method\_metrics\_set_{me, a} = me_{aggr}(a,i), \forall i\in I$. These aggregate values are used to give a better perspective on the mean differences of importance in image regions between augmented and baseline model CAMs, as interpreted by the metric $me$.
    \item The last step consists of using the selected methods to generate a quantitative analysis of the metrics to determine the impacts each augmentation has when compared to the baseline model, and then going over the generated analysis.
\end{enumerate}

This methodology, as previously mentioned, is based on \citep{tang2020explaining}, but here we provide a structured set of steps that applies to different architectures, datasets, metrics, augmentation techniques, and CAM generation methods. For the exploration made in this work, we also provide a richer set of metrics between CAMs, each of them described in the following Section (\hyperref[ch:experiments]{Section \ref*{ch:experiments}}).

\section{Experiments}\label{ch:experiments}

In this section, we will discuss how we applied the methodology, delineating our choice of augmentations, metrics, and datasets in \hyperref[ch:experiments:design]{Section \ref*{ch:experiments:design}}, as well as discussing our choices and showing the results we obtained using it in \hyperref[ch:experiments:results]{Section \ref*{ch:experiments:results}}.

\subsection{Experiment Design}\label{ch:experiments:design}

In the first part of this section, we discuss our choices for the dataset and base architecture. After that, we will expand upon important details concerning how datasets were pre-processed and how training was conducted. Lastly, we will discuss the metrics we used between CAMs, and briefly examine other important details of our application.

When considering our choice of dataset for this experiment, we found it important to find one that struck a balance between complexity, to allow for rich analysis, and training time. Considering this, we used CIFAR-10 \citep{krizhevsky2009learning}, an image classification dataset with 60,000 images. It has properties that favor a richer analysis and more complex challenge than the MNIST dataset used by \citep{tang2020explaining}, and is also commonly used in the literature \citep{inoue2018data, cubuk2019autoaugment, ratner2017learning, liu2015very, hussain2019study}. CIFAR-10 has a relatively small set of classes at 10 labels, colored images representing relatively complex and varied patterns, a balanced distribution of samples in each class, and relatively low image resolution, making it an adequately challenging dataset for our image classification purposes.

\begin{figure}[H]
\caption{Evolution of the accuracy on the test set across 30 training epochs for the three candidate architectures.}
\label{fig:arq:accuracy}
\includegraphics[width=.5\linewidth]{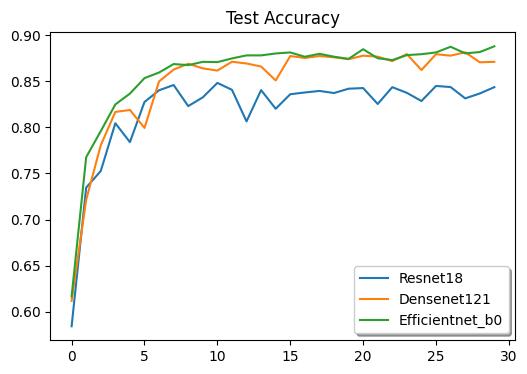}
\centering
\end{figure}

\begin{figure}[H]
\caption{Evolution of the precision on the test set across 30 training epochs for the three candidate architectures.}
\label{fig:arq:precision}
\includegraphics[width=.5\linewidth]{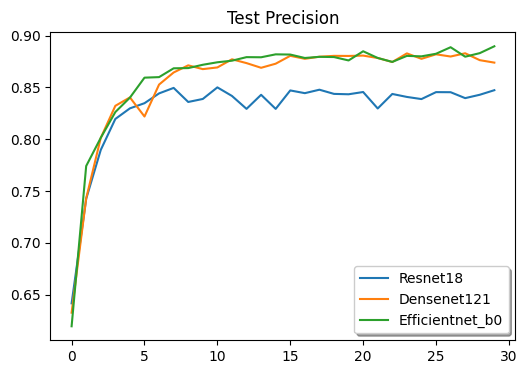}
\centering
\end{figure}

\begin{figure}[H]
\caption{Evolution of the recall on the test set across 30 training epochs for the three candidate architectures.}
\label{fig:arq:recall}
\includegraphics[width=.5\linewidth]{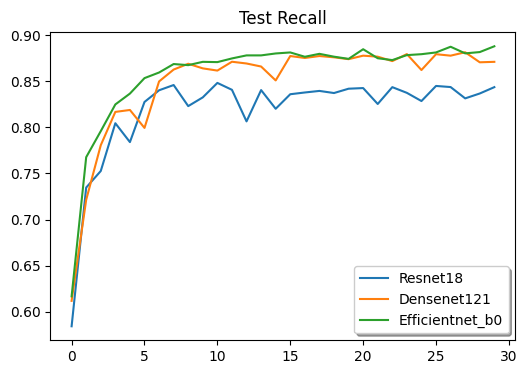}
\centering
\end{figure}

\begin{figure}[H]
\caption{Evolution of the F1-score on the test set across 30 training epochs for the three candidate architectures.}
\label{fig:arq:f1_score}
\includegraphics[width=.5\linewidth]{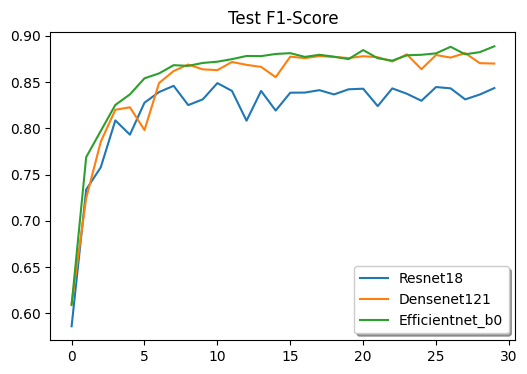}
\centering
\end{figure}

\begin{figure}[H]
\caption{Comparison of the amount of time necessary to train each of the candidate architectures over 30 epochs.}
\label{fig:arq:time}
\includegraphics[width=.5\linewidth]{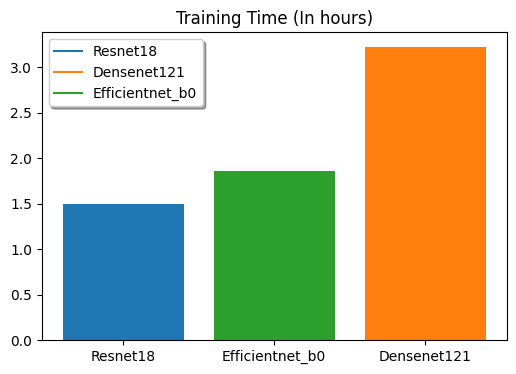}
\centering
\end{figure}

To determine which base architecture to use, we selected three commonly used architectures in the literature (EfficientNet B0 \citep{tan2019efficientnet}, ResNet-18 \citep{he2016deep}, DenseNet-121 \citep{huang2017densely}) and measured their performances in image classification with CIFAR-10. To compare them, we ran training routines for the three architectures for 30 epochs over the CIFAR-10 dataset and measured their test accuracy, precision, F1-score, and recall for each epoch. After obtaining the results for the model trainings (\autoref{fig:arq:accuracy} for accuracy, \autoref{fig:arq:precision} for precision, \autoref{fig:arq:recall} for recall, \autoref{fig:arq:f1_score} for F1-score, and \autoref{fig:arq:time} for training time), we chose EfficientNet B0 for its balance between training time and performance.

As mentioned in \hyperref[ch:bg:da]{Section \ref*{ch:bg:da}}, we chose seven data augmentation methods that we felt best represented a wide spread of different augmentation techniques, which would provide a richer ground for data augmentation analysis. The methods we chose were affine transformations, cutmix, color jitter, equalization, elastic transforms, random cropping, and Gaussian blurring.

For the train-validation-test split, as CIFAR10 comes with a train-test split dataset, with 50,000 images for training and 10,000 images for testing, the only further modification we needed was a train-validation split. For that purpose, we decided on using random stratified sampling with 90\% train to 10\% validation split, with the same 45,000 base images for training and 5,000 for validation in all models. We do not use the validation set for hyperparameter optimization. We instead use the validation set to select the version of the model with the highest accuracy against it across all epochs. For the augmented datasets, we generated one augmented image for each training image, resulting in 90,000 total images for each augmented dataset. These augmentations are done online during batch fetching. For the training process, we adopted the cross-entropy loss as our loss function and adopted the Adam optimizer, with a learning rate at $1e^-3$ and weight decay at $1e-5$. For training, validation, and testing routines, we used batch sizes of 32, and trained all models for 30 epochs each.

To control the impact that the order of image batches could have during training, it was important to manage, across all augmented datasets, the order in which images are fed to the models during training. We discuss more of how this was achieved in \hyperref[ch:appendix:determinism]{Appendix \ref*{ch:appendix:determinism}}. The core idea was to guarantee, as much as possible, that the placement order for all augmented and unaugmented images in a dataset would be the same across all augmented datasets. For the unaugmented images, following this requirement meant placing them in the same order across augmented datasets. For the augmented images, we met this requirement through a different method. Instead of guaranteeing that the same augmented image appears in the same ordering position across datasets (which would be impossible as each combination of augmentation method and starting state applied over the same source image produces a different augmented image,) we instead use the source of the augmented images to order them. We guarantee that the source image for an augmented image at a specific position in the dataset will be the same across datasets, and do the same for all augmented images.

Additionally, regardless of augmentation,  all images were resized to $224\times224$ pixels (while preserving the aspect ratio) to fit the input format of EfficientNet B0 and normalized in order to improve results. To normalize the dataset, we calculate the mean and standard deviation of each color channel across all images in the dataset. From the mean and standard deviation, we alter the value of each RGB channel of a pixel by subtracting the respective color channel's mean from the pixel value of the channel and then dividing the result by that color channel's standard deviation. By doing this, every pixel's color channel value will have a mean of 0 and a standard deviation of 1 across the entire dataset. This significantly reduces the variability range of the dataset, accelerating the training convergence time.

Another important detail to cover is how we define the starting states $X$. Here, we choose three different states $X=[Seed 1, Seed 2, Seed 3]$ for each augmentation, with each state representing a unique seed fed to all random generation libraries involved in the training of the respective augmented model. It is important to note that these states are stable across all augmentations, meaning that the seed used as the starting state for $M_{RandCrop, Seed 1}$ is the same as that of $M_{Cutmix, Seed 1}$, which is the same as $M_{Affine, Seed 1}$, and so on.

For the CAM generation, as detailed previously, we chose the default Grad-CAM algorithm. In order to generate CAMs with this algorithm, we must provide a target class alongside a model, the target layer of said model, and the input image. This is necessary as the CAM is a measure of which areas of an image the CNN model is the most sensitive to with regard to a specific class. Another interpretation of CAMs is that regions with high values are regions where a change in image pixel values would significantly impact a model's output with regard to the given class. For the choice of target label, we used the model's predicted class for the respective test image. An alternative to this choice would've been to use the ground truth label instead of the predicted class, which would show us what the CAM would look like for a given label, even if the model itself would not have made that classification choice. The problem with this lies with the usage of a CAM with a target class that the respective model would not have predicted, which would mean we are not analyzing the final decision-making process of the model representative of the patterns it has learned, but instead how the model would act if it had perfect performance. The upside to that approach, however, would be that the baseline and augmented model CAMs would always have the same class, and thus would theoretically make analysis easier and more straightforward. Both approaches have their own merits, and it is not clear which of them is the ideal approach. Due to our interest in analyzing the behavioral patterns ultimately learned by the models, we decided to use the predicted label instead of the ground truth as the target class for CAM generation, even if the behavioral patterns they exhibit involve wrong classifications or classifications that differ between baseline and augmented models.

To evaluate the differences between the CAMs generated by the baseline and those generated by the different augmented models, we used the following metrics between CAMs: Mean Absolute Difference (MAD), Mean Squared Difference (MSD), Pearson Correlation, Spearman Correlation, and Overlap Rate. Additionally, we also use Kullback-Leibler Divergence (KLD) between the class predictions for each image, which here we will call Class Prediction Kullback-Leibler Divergence (Class-KLD). The discrete probability distributions used to calculate Class-KLD are explained ahead in \hyperlink{ch:experiments:design:kld}{Section 5.1}. Each metric is explained as follows:


\begin{description}
    \item [Mean Absolute Difference (MAD):]  It is the mean absolute error (MAE) metric applied between two CAMs. We reinterpret this error that MAE measures as the mean pixel difference between the two CAMs. With this metric, we are purely interested in the magnitude of the difference between the maps, so this reinterpretation does not introduce any unintended problems in analyzing this metric. Although this metric has no specific unit, considering the range of values $[0, 1]$ a pixel can take, the maximum mean difference is $1$, indicating that baseline and augment images have the maximum possible difference, and the minimum is $0$, indicating two identical maps. This metric may be interpreted as a measure of the mean magnitude of the differences between a pair of CAMs, and can also be viewed as a degree of dissimilarity. Considering a pair of CAMs $P$ and $Q$, and pixel indexes $j$ for source image $J$ with the total number of pixels $N=|J|$, where $P_j$ represents the activation of pixel $j$ in the CAM P, MAD is defined as
    \begin{center}
        $MAD(P, Q) = \frac{1}{N}\sum\limits_{j}|P_j - Q_j|$
    \end{center}

    \item [Mean Squared Difference (MSD):] It follows the same logic as MSD, but squaring the difference instead of taking the absolute of the difference. We use this metric in case it may exacerbate large differences between CAMs and aid in analyzing the magnitudes of differences from another perspective. MSD is defined as
    \begin{center}
        $MSD(P, Q) = \frac{1}{N}\sum\limits_{j}(P_j - Q_j)^2$
    \end{center}

    \item [Pearson Correlation Coefficient:] Pearson Correlation is a metric used to measure the linear relationship between two variables, reflecting the strength and direction of this relationship \citep{bylinskii2018different}. It can be understood as a measure of the degree to which two variables covary while maintaining a reasonably constant proportion. To use it, we flatten both CAM images into 1D vectors and then measure their covariance, using the baseline as $P$ and augmented as $Q$. It is defined as 
    \begin{center}
        $Pearson(P, Q) = \frac{cov(P, Q)}{\sigma(P) \times \sigma(Q)}$
    \end{center}
    where $cov(P,Q)$ is the covariance between $P$ and $Q$, and $\sigma(P)$ and $\sigma(Q)$ are the standard deviation of $P$ and $Q$, respectively.

    \item [Spearman Correlation:] Spearman follows in the same steps as Pearson, but as this metric weighs monotonicity more heavily than proportionality. That is, it measures how well the relationship between two variables can be described using a monotonic (but not necessarily linear) function. Alongside Pearson, it helps paint a clearer picture of the way baseline and augmentation covary. It is defined as
    \begin{center}
        $Spearman(P, Q) = \frac{cov(R(P), R(Q))}{\sigma(R(P)) \times \sigma(R(Q))}$
    \end{center}
    where $R(P)$ and $R(Q)$ refer to the rank-variable versions of $P$ and $Q$, meaning $P$ and $Q$ transformed into a weak-ordering vector where any two items necessarily have a 'higher than', 'lower than' or 'equal to' relationship.

    \item [Overlap Rate:] It is a specialization of the metric Intersection over Union \citep{rezatofighi2019generalized} as applied to a specific selection of pixel regions. Intersection over Union works by analyzing two regions of an image and taking a ratio of the intersection between these regions over the union between these regions \citep{rezatofighi2019generalized}. Overlap Rate follows the same idea, but selects the region automatically by using the $Y\%$ pixels with the highest activations in the CAM, meaning the pixels with the highest value. This gives us an idea of how the important areas differ in spatial distribution, which will complement the information gained from MSD and MAD regarding the magnitude of their difference. It is important to note, also, that the percentage is not referring to a threshold of $Y\%$ of population, but rather to a threshold based on the value of the $Y$-th population percentile; this means that if $Y=10$ and the 10th percentile highest value is $0.7$, we will select all pixels with threshold value equal to or higher than $0.7$. This may result in selecting more than 10\% of the population if there are multiple values equal to $0.7$ in the population.

    \item [Class Prediction Kullback-Leibler Divergence:] \hypertarget{ch:experiments:design:kld} Class-KLD follows a different idea from the others. While the other metrics are applied directly between the CAMs, we apply Kullback-Leibler Divergence to the prediction matrix made by model $M_{a,x}$ as compared to the prediction made by the baseline model $B$. KLD is defined as an information-theoretic measure of the difference between two probability distributions \citep{bylinskii2018different}. The idea behind the use of this metric is to measure how different the predictions made by the two models are for a test image. Between class prediction vectors P and Q, KLD is defined as
    \begin{center}
        $KLD(P, Q) = \sum_{i} Q_{i} log(\epsilon + \frac{Q_{i}}{P_{i}})$
    \end{center}
    where, in our application, $i$ refers to class in the vector of predictions, and $\epsilon$ is a regularization constant.
\end{description}

Having defined the base architecture, dataset, augmentations, and metrics, the application of the methodology is straightforward. First, we train the baseline and augmented models and then use Grad-CAM to generate CAM sets for all the models. From there, we collect the metrics between each augmented model's CAM set and the baseline model $B$'s CAM set, considering all the images in the test set. After averaging the $|X|$ models' metrics for each augmentation, we then analyze these metrics.

To analyze the results we obtained, we adopted the following approaches:
\begin{itemize}
    \item Analysis of performance metrics between the models. Although this is not the main focus of our work, determining their differences in effectiveness at the classification task is relevant information.
    \item Analysis of the similarity metrics value distribution over the test image set for every augmentation. Using boxplots, we inspect properties of the metric value distributions of the augmentations, which may be able to establish trends in the collected data. This analysis, however, is aggregative in nature; it does not permit an individual investigation into how each augmentation behaves for each test image.
    \item Correlation analysis for each metric across augmentations. By correlating the metric values across augmentations, we view another aspect of the behaviors between augmentations. For each metric $me\in ME$, we take the average metric sets $aug\_method\_metrics\_set_{me, a}$ for all augmentations $a\in A$ and analyze how these sets of metrics correlate with one another across augmentations. This analysis enables us to understand how the metrics covary between augmentations across the test image set, allowing us to make more nuanced observations about changes in behavior induced by the augmented dataset.
\end{itemize}


From these analyses, our objective is to build a comprehensive view of how the different models behave. The metric distribution plots, performance metric comparisons, and correlation maps are shown in \hyperref[ch:experiments:results]{Section \ref*{ch:experiments:results}}, and the results as a whole are analyzed in \hyperref[ch:experiments:results:analysis]{Section \ref*{ch:experiments:results:analysis}}.

\subsection{Results}\label{ch:experiments:results}

Below are the metric results taken between the baseline and augmented models. Firstly, we will analyze performance metrics related to the augmentations. Secondly, we will show the boxplot and correlation map results for every metric, analyzing them individually. Afterward, we will inspect the results more generally, making observations on trends we see and interpreting the results.

\subsubsection{Performance Metrics}

\begin{figure}[H]
\caption{Bar plot of the accuracy performance metric for all models over the test image set.}
\label{fig:bar:accuracy}
\includegraphics[width=.6\linewidth]{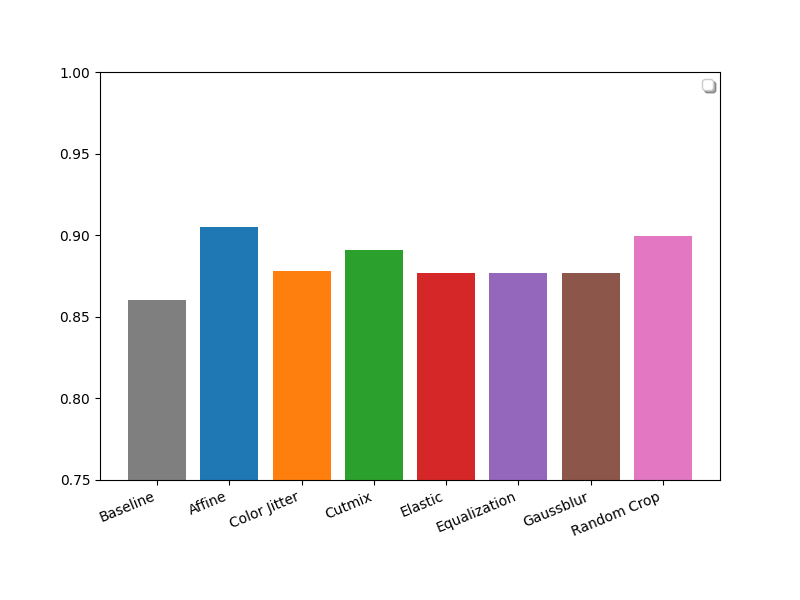}
\centering
\end{figure}

\begin{figure}[H]
\caption{Bar plot of the precision performance metric for all models over the test image set.}
\label{fig:bar:precision}
\includegraphics[width=.6\linewidth]{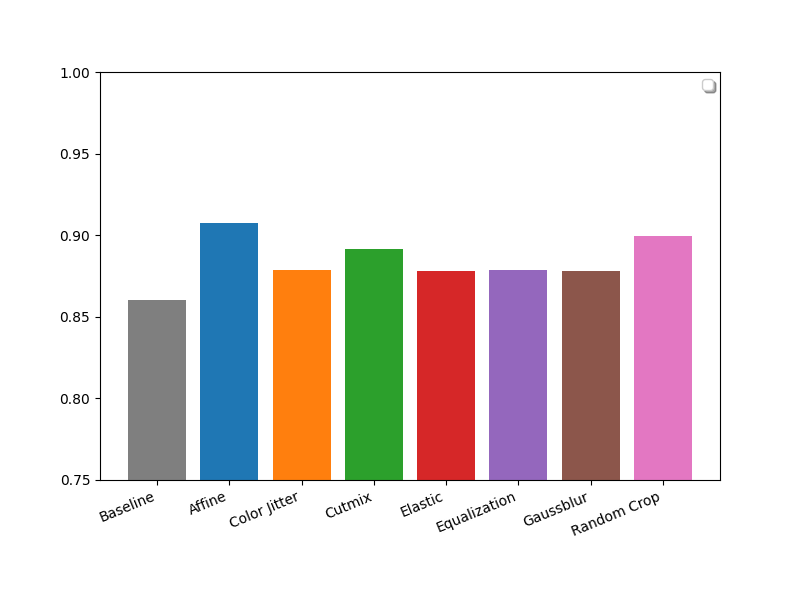}
\centering
\end{figure}

\begin{figure}[H]
\caption{Bar plot of the recall performance metric for all models over the test image set.}
\label{fig:bar:recall}
\includegraphics[width=.6\linewidth]{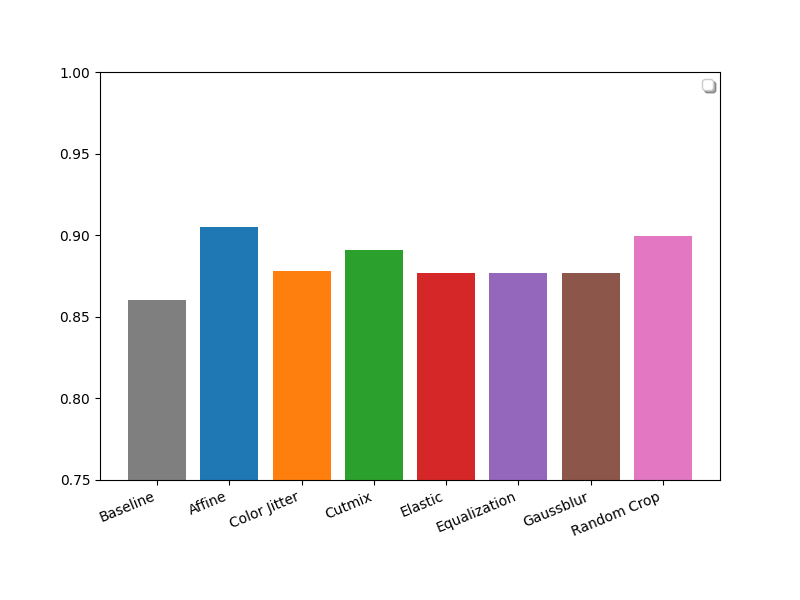}
\centering
\end{figure}

\begin{figure}[H]
\caption{Bar plot of the F1-score performance metric for all models over the test image set.}
\label{fig:bar:f1_score}
\includegraphics[width=.6\linewidth]{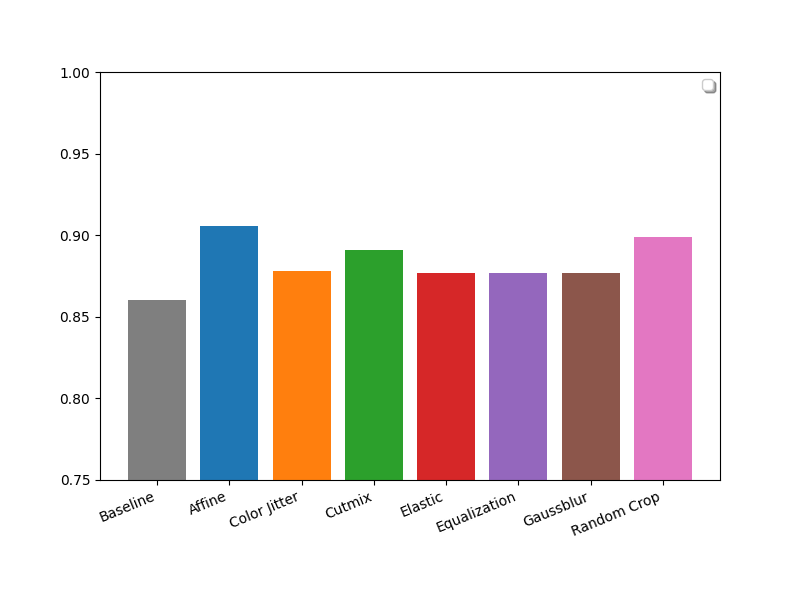}
\centering
\end{figure}

To measure model performance,  we tracked, as previously mentioned, the accuracy, F1-measure, recall, and precision of all the models. Below are the results of these metrics over the test set for all augmentations, with the mean performance of the $|X|$ models being shown for each augmentation. \autoref{fig:bar:accuracy} shows the average test accuracy between the models, \autoref{fig:bar:precision} shows the average test precision, \autoref{fig:bar:recall} shows the recall comparison, and \autoref{fig:bar:f1_score} shows the F1-score comparison.

The results are consistent across metrics, implying that none of the models suffer particularly from the impact of false positives or false negatives. Additionally, although not an unexpected result, we note that utilizing any of the seven augmentation methods results in increased performance over the baseline model. All models display similar performance to one another, with affine transforms achieving the best performance across all performance metrics. 

Also, as expected, it is hard to draw any conclusions about the impact of data augmentation by inspecting only performance metrics; to that end, we will have to analyze the CAM metrics. In the following sections, we will inspect these metrics individually.

\subsubsection{Overlap Rate (Top 20, Top 10, Top 5)}

\begin{figure}[H]
\caption{Boxplot for overlap rate with threshold $Y=20$, showing the distribution of the metric values over the test set for each augmentation.}
\label{fig:box:overlap_20}
\includegraphics[width=.6\linewidth]{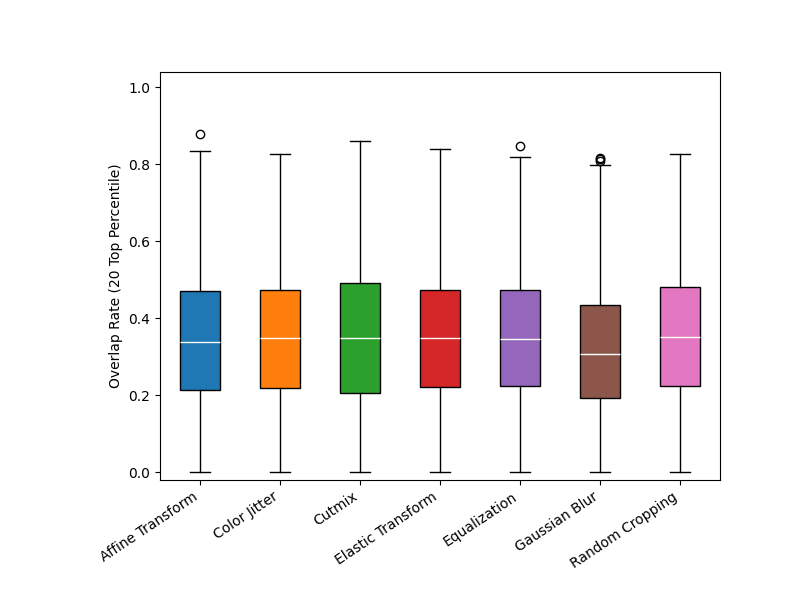}
\centering
\end{figure}

\begin{figure}[H]
\caption{Correlation map for overlap rate with threshold $Y=20$, showing the correlation of the metric between augmentations over the test set.}
\label{fig:cor:overlap_20}
\includegraphics[width=.6\linewidth]{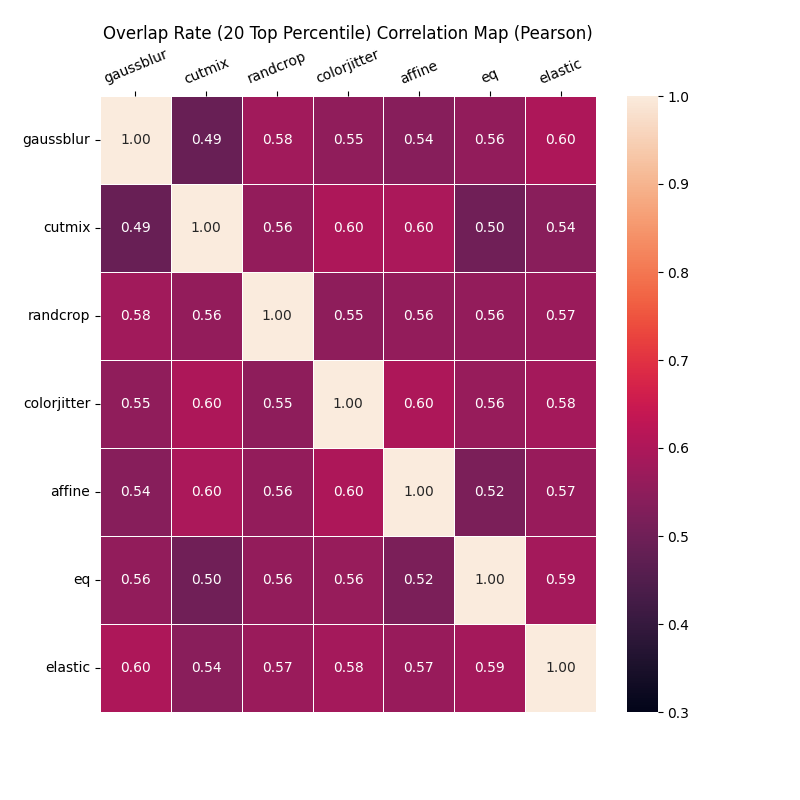}
\centering
\end{figure}

\begin{figure}[H]
\caption{Boxplot for overlap rate with threshold $Y=10$, showing the distribution of the metric values over the test set for each augmentation.}
\label{fig:box:overlap_10}
\includegraphics[width=.6\linewidth]{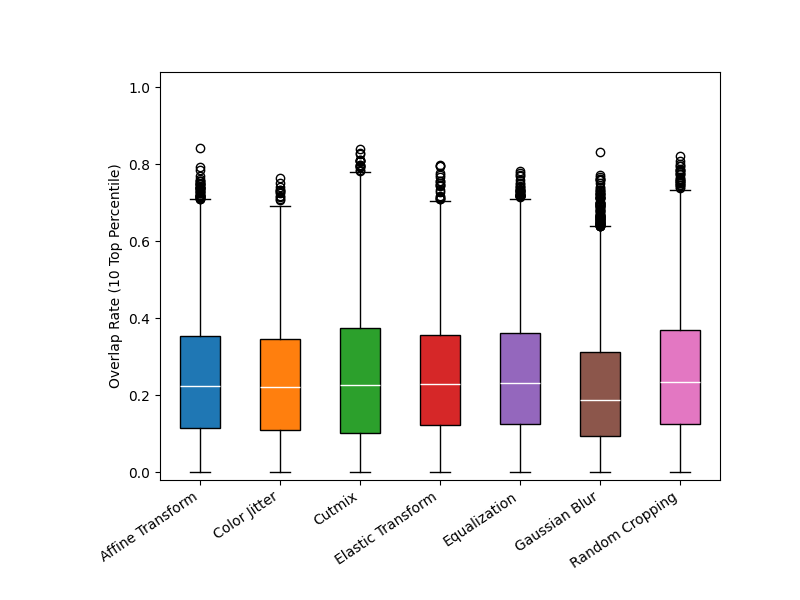}
\centering
\end{figure}

\begin{figure}[H]
\caption{Correlation map for overlap rate with threshold $Y=10$, showing the correlation of the metric between augmentations over the test set.}
\label{fig:cor:overlap_10}
\includegraphics[width=.6\linewidth]{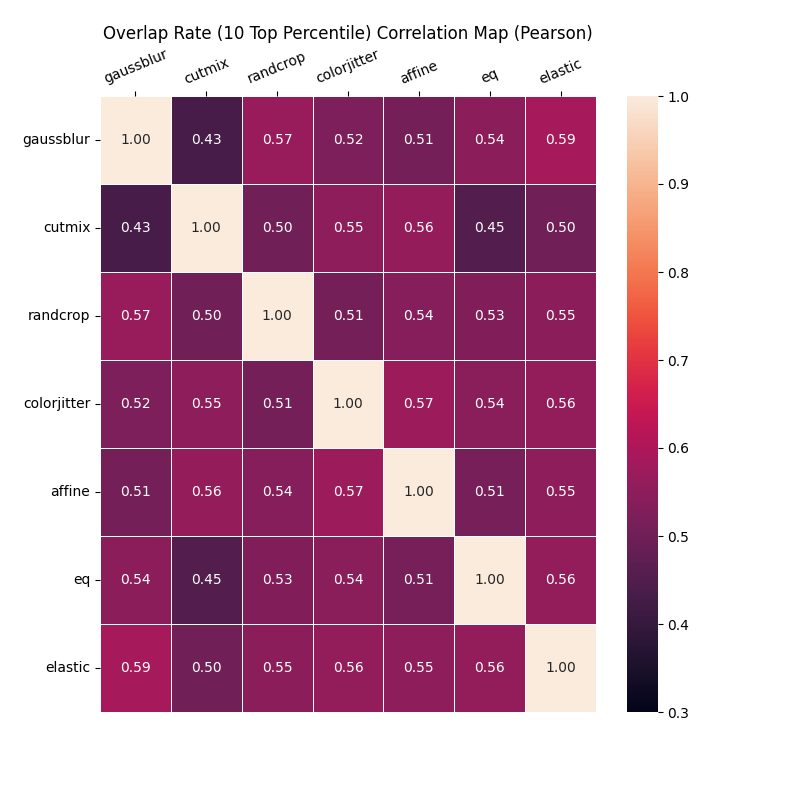}
\centering
\end{figure}

\begin{figure}[H]
\caption{Boxplot for overlap rate with threshold $Y=5$, showing the distribution of the metric values over the test set for each augmentation.}
\label{fig:box:overlap_5}
\includegraphics[width=.6\linewidth]{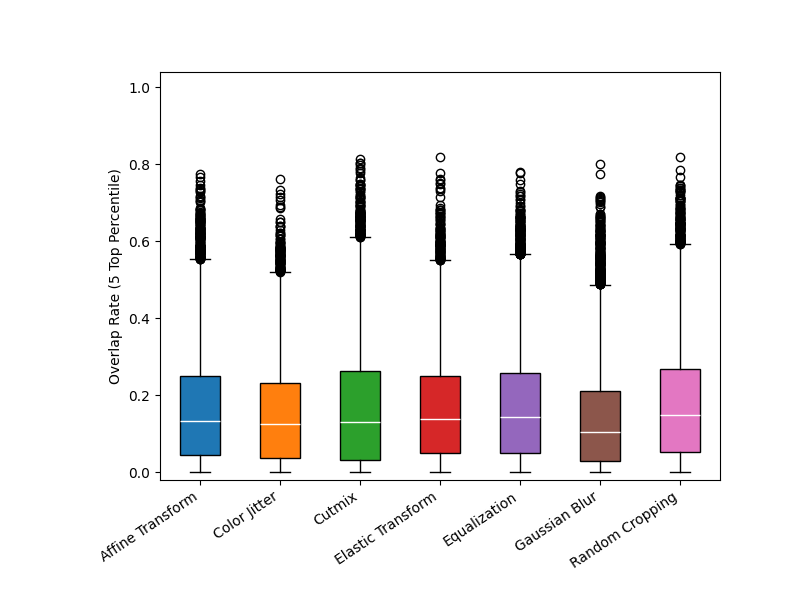}
\centering
\end{figure}

\begin{figure}[H]
\caption{Correlation map for overlap rate with threshold $Y=5$, showing the correlation of the metric between augmentations over the test set.}
\label{fig:cor:overlap_5}
\includegraphics[width=.6\linewidth]{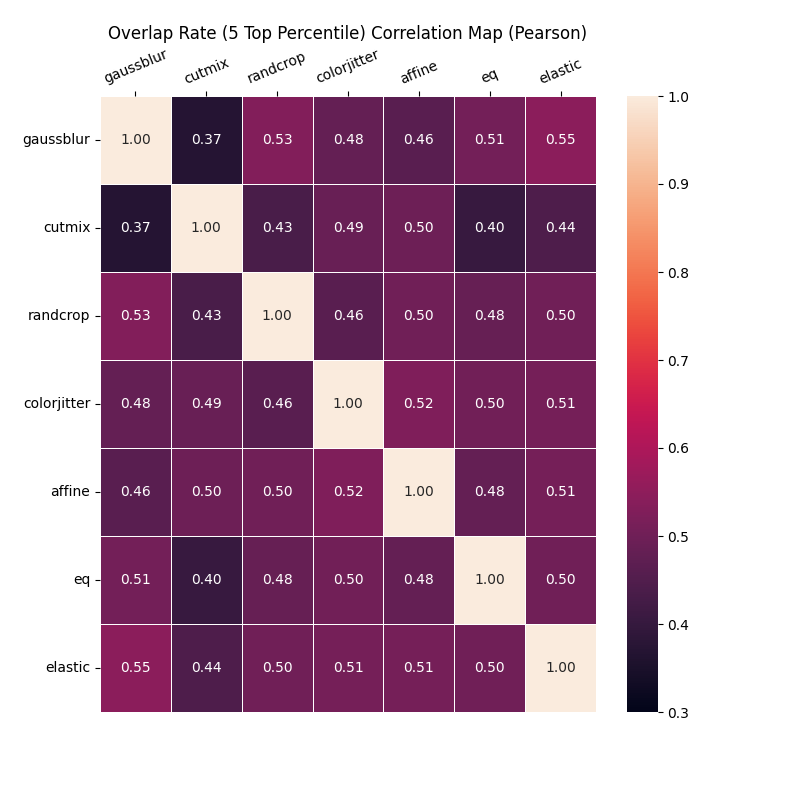}
\centering
\end{figure}

In this section, we will show boxplot and correlation map results for the overlap rate metric. In \autoref{fig:box:overlap_20}, we show the results for overlap rate with a threshold of $Y=20$, and in \autoref{fig:cor:overlap_20} we show the correlation map across augmentations for overlap rate with $Y=20$. We do the same for $Y=10$ (\autoref{fig:box:overlap_10} for the boxplot, \autoref{fig:cor:overlap_10} for the correlation map) and for $Y=5$ (\autoref{fig:box:overlap_5} showing the boxplot, \autoref{fig:cor:overlap_5} showing the correlation map).

Analyzing the differences between overlap rate boxplots, we can see that as the importance threshold increases (and $Y$ lowers from 20 (\autoref{fig:box:overlap_20}) to 10 (\autoref{fig:box:overlap_10}) to 5 (\autoref{fig:box:overlap_5})), the observed overlap rate lowers for all augmentations. This behavior indicates that, although the augmented models may consider similar image regions to be similarly important overall, they focus on different parts of the image as we filter out the less important regions of the CAMs. This result is not necessarily unexpected, but it is worth noting as it quantitatively confirms that when considering the most significant minutiae of an image, augmented models consider different image regions as highly important towards their classification compared to the baseline.

When compared to the Top 20\% correlation map (\autoref{fig:cor:overlap_20}), we can also see that the Top 5\% overlap rate correlation map (\autoref{fig:cor:overlap_5}) has lower correlation values across augmentations, with the majority of them being at or below 0.5. This observation may indicate that not only do augmented models look at different regions than the baseline model, but how these important regions overlap with the baseline may also differ between augmentations. This observation is supported by the fact that if all augmented model CAMs were to, on average, differ from the baseline ones in the same ways, we could expect their correlation levels to be significantly higher, which does not happen. This observation could be made qualitatively by inspecting many different samples of CAMs, but through this method, we can confirm that this analysis holds throughout the entirety of the dataset in a quantitative manner without the need for a human to compare the maps individually.
\subsubsection{Pearson and Spearman correlations}

\begin{figure}[ht]
\caption{Boxplot for Pearson Correlation Coefficient, showing the distribution of the metric values over the test set for each augmentation.}
\label{fig:box:pearson}
\includegraphics[width=.6\linewidth]{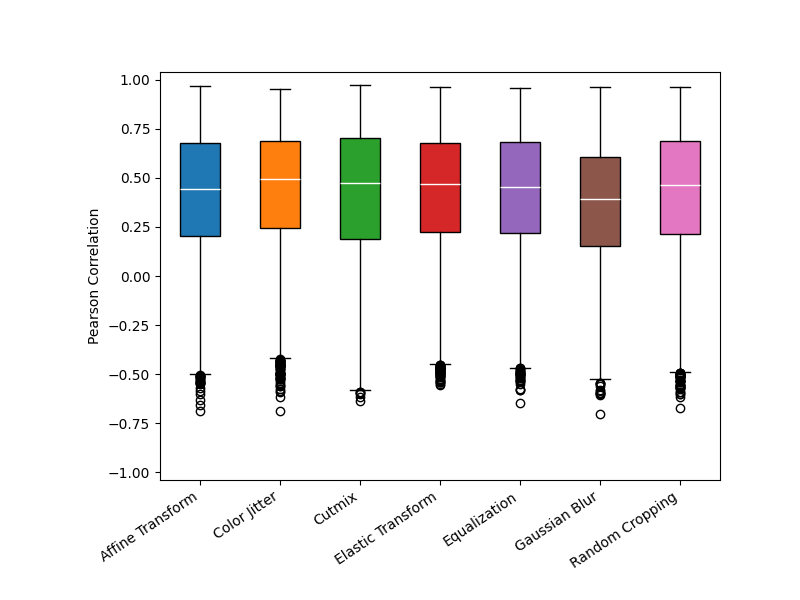}
\centering
\end{figure}

\begin{figure}[ht]
\caption{Correlation map for Pearson Correlation Coefficient, showing the correlation of the metric between augmentations over the test set.}
\label{fig:cor:pearson}
\includegraphics[width=.6\linewidth]{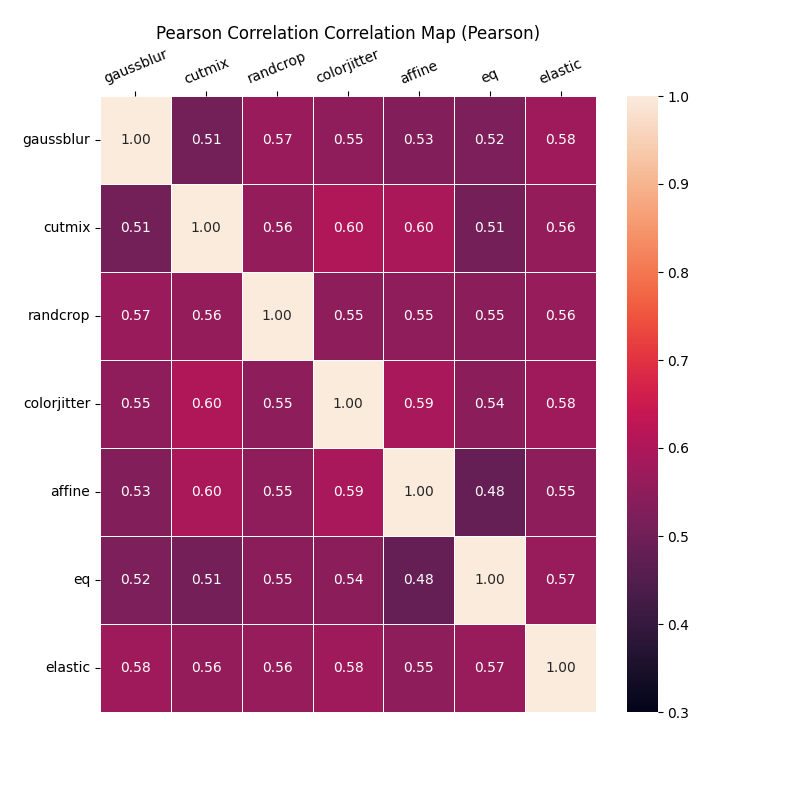}
\centering
\end{figure}

\begin{figure}[ht]
\caption{Boxplot for Spearman Correlation Coefficient, showing the distribution of the metric values over the test set for each augmentation.}
\label{fig:box:spearman}
\includegraphics[width=.6\linewidth]{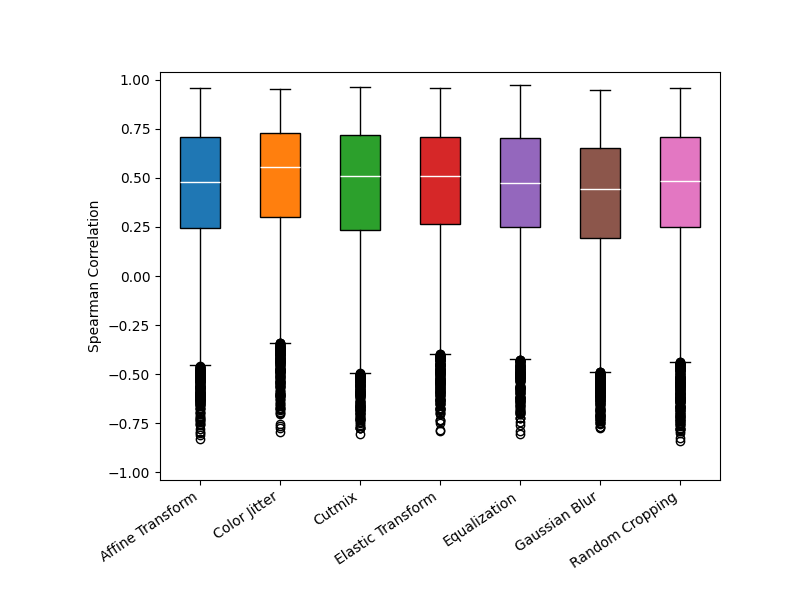}
\centering
\end{figure}

\begin{figure}[H]
\caption{Correlation map for Spearman Correlation Coefficient, showing the correlation of the metric between augmentations over the test set.}
\label{fig:cor:spearman}
\includegraphics[width=.6\linewidth]{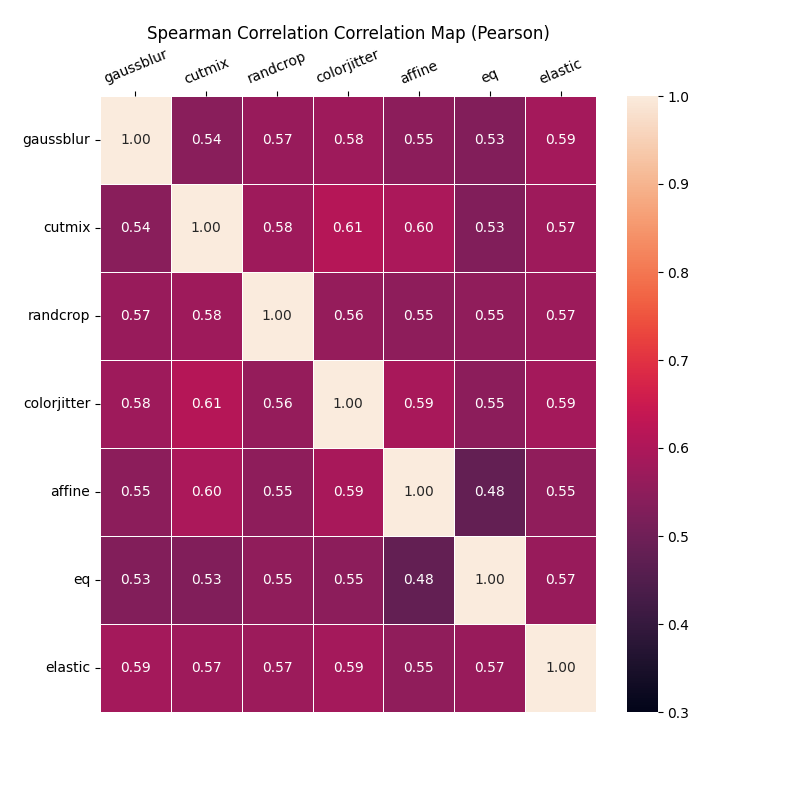}
\centering
\end{figure}

In this section, we will present and discuss boxplot and correlation map results across augmentations for the Pearson and Spearman correlation metrics. \autoref{fig:box:pearson} shows the boxplot for Pearson correlation, and \autoref{fig:cor:pearson} shows the correlation map for the metric across augmentations. \autoref{fig:box:spearman} shows the distribution plot for Spearman correlation, and \autoref{fig:cor:spearman} shows its correlation map.

\raggedbottom
For the Spearman and Pearson metric distributions (\autoref{fig:box:spearman}, \autoref{fig:box:pearson}), we notice a moderate correlation between the baseline and augmented CAMs. Note that while a large portion of the population presents correlations between baseline and augmented CAMs above 0.5 for both the Pearson (\autoref{fig:box:pearson}) and Spearman (\autoref{fig:box:spearman}) boxplots, the median line for all augmentations is below that value, indicating only moderate correlation. We also note that a significant amount of augmented model CAMs have an inverse correlation to baseline CAMs, which indicates that, for those images, when the activation for a region rises in baseline, the activation for that region lowers in the augmented model, and vice versa. One notable observation from the correlation map for Pearson across augmentations (\autoref{fig:cor:pearson}) is that all augmentations have lower Pearson correlation metric values across augmentations on average compared to their Spearman correlation metrics (\autoref{fig:cor:spearman}). This can also be evidenced in the, on average, lower values between the Pearson metric distributions (\autoref{fig:box:pearson}) and their Spearman metric counterparts (\autoref{fig:box:spearman}). This observation indicates that the correlation these augmented model CAMs have to the baseline is more alike in monotonicity than in terms of linear relationship.

\subsubsection{Mean Absolute Difference and Mean Squared Difference}
\begin{figure}[H]
\caption{Boxplot for Mean Absolute Difference, showing the distribution of the metric values over the test set for each augmentation.}
\label{fig:box:mae}
\includegraphics[width=.6\linewidth]{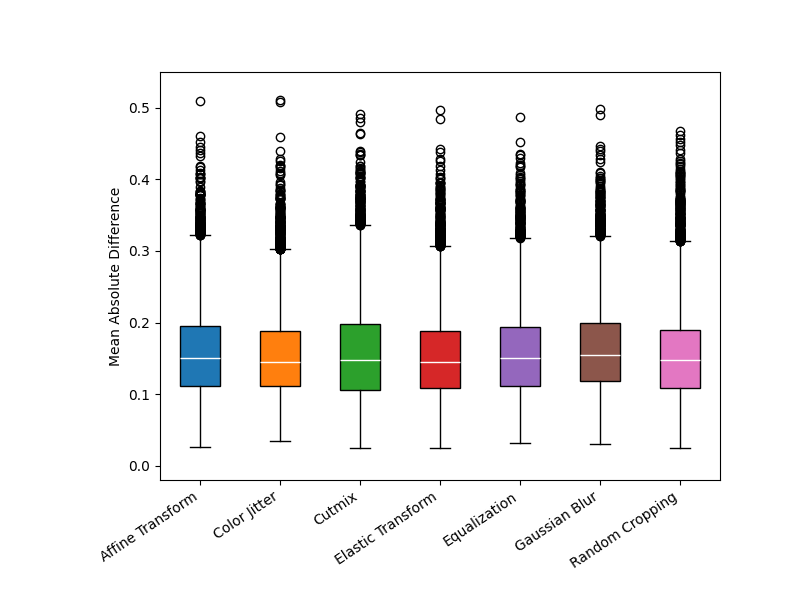}
\centering
\end{figure}

\begin{figure}[H]
\caption{Correlation map for Mean Absolute Difference, showing the correlation of the metric between augmentations over the test set.}
\label{fig:cor:mae}
\includegraphics[width=.6\linewidth]{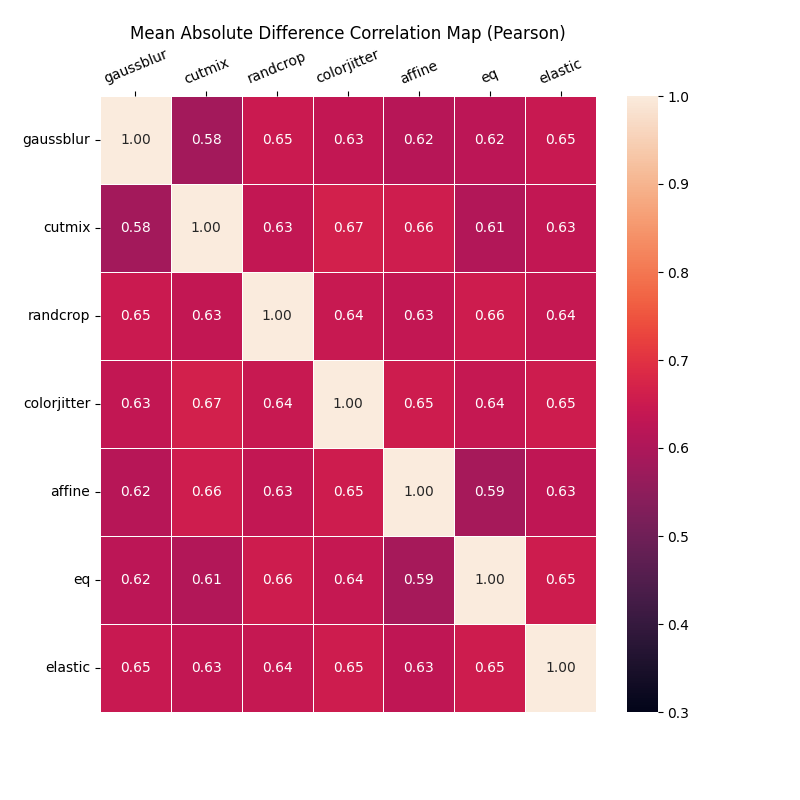}
\centering
\end{figure}

\begin{figure}[H]
\caption{Boxplot for Mean Squared Difference, showing the distribution of the metric values over the test set for each augmentation.}
\label{fig:box:mse}
\includegraphics[width=.6\linewidth]{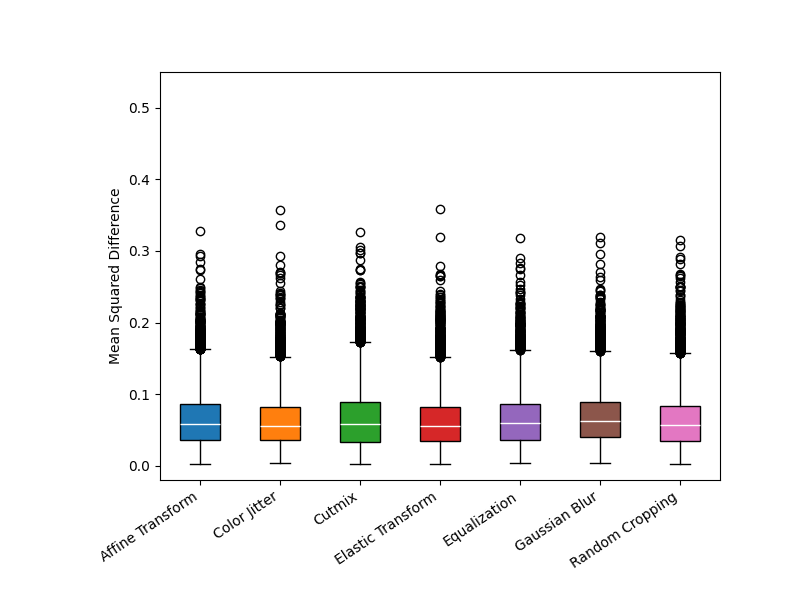}
\centering
\end{figure}

\begin{figure}[H]
\caption{Correlation map for Mean Squared Difference, showing the correlation of the metric between augmentations over the test set.}
\label{fig:cor:mse}
\includegraphics[width=.6\linewidth]{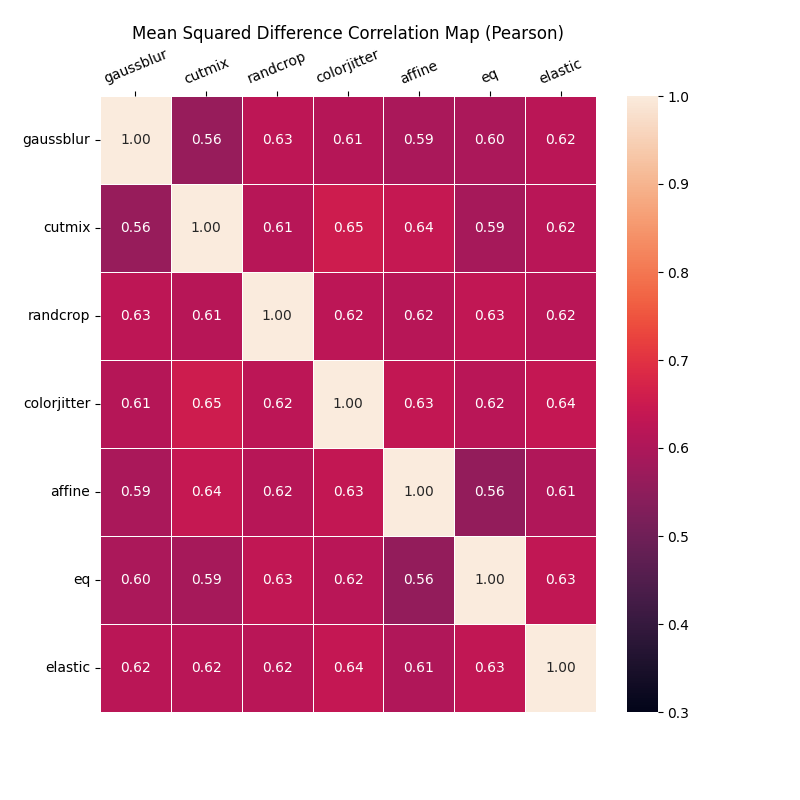}
\centering
\end{figure}

In this section, we share the results for the MAD and MSD metrics. \autoref{fig:box:mae} shows the boxplot of MAD, while \autoref{fig:cor:mae} shows the correlation map for MAD. For MSD, \autoref{fig:box:mse} shows its boxplot and \autoref{fig:cor:mse} shows its correlation map.

We can discern from the MSD (\autoref{fig:box:mse}) and MAD (\autoref{fig:box:mae}) boxplots that when the CAMs differ between baseline and augmentation, the average magnitude of the pixel value difference follows similar trends throughout augmentations. None of the augmentations shows a significant increase or decrease in their metric distributions compared to the others. One observation to note is that all the augmentation pairs in the correlation maps across augmentations (\autoref{fig:cor:mae} for MAD, \autoref{fig:cor:mse} for MSD) present moderate correlations (around 0.5 and 0.6), which implies that the magnitude of difference between CAM and baseline maps is correlated throughout the augmentations. Considering both observations, they indicate that, in general, the mean magnitude of pixel activation differences between baseline and augmented model CAMs is not high across all augmentations for each test image. Moreover, considering these two metrics for each image, we note that these values covary positively and with moderate force between augmentation pairs.

\subsubsection{Class Prediction Kullback-Leibler Divergence}

\begin{figure}[H]
\caption{Boxplot for Class-KLD, showing the distribution of the metric values over the test set for each augmentation.}
\label{fig:box:class_kld}
\includegraphics[width=.6\linewidth]{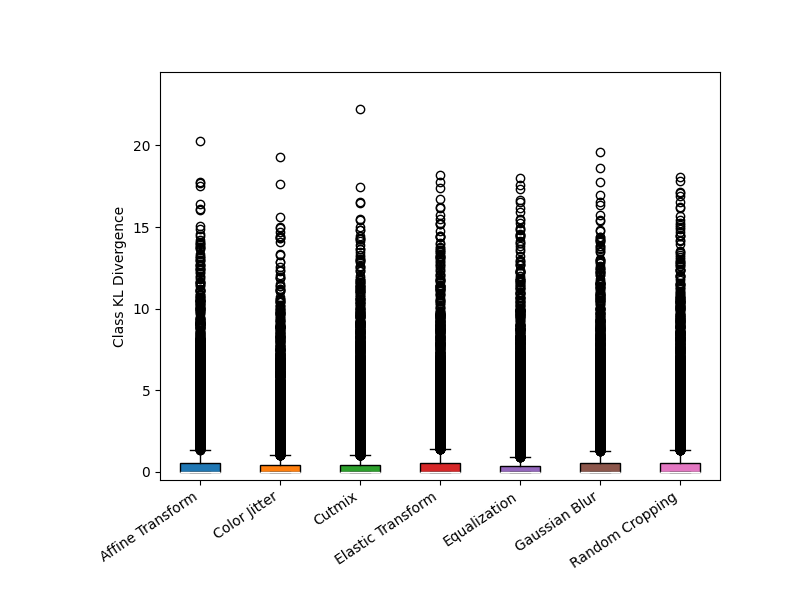}
\centering
\end{figure}

\begin{figure}[ht]
\caption{Correlation map for Class-KLD, showing the correlation of the metric between augmentations over the test set.}
\label{fig:cor:class_kld}
\includegraphics[width=.6\linewidth]{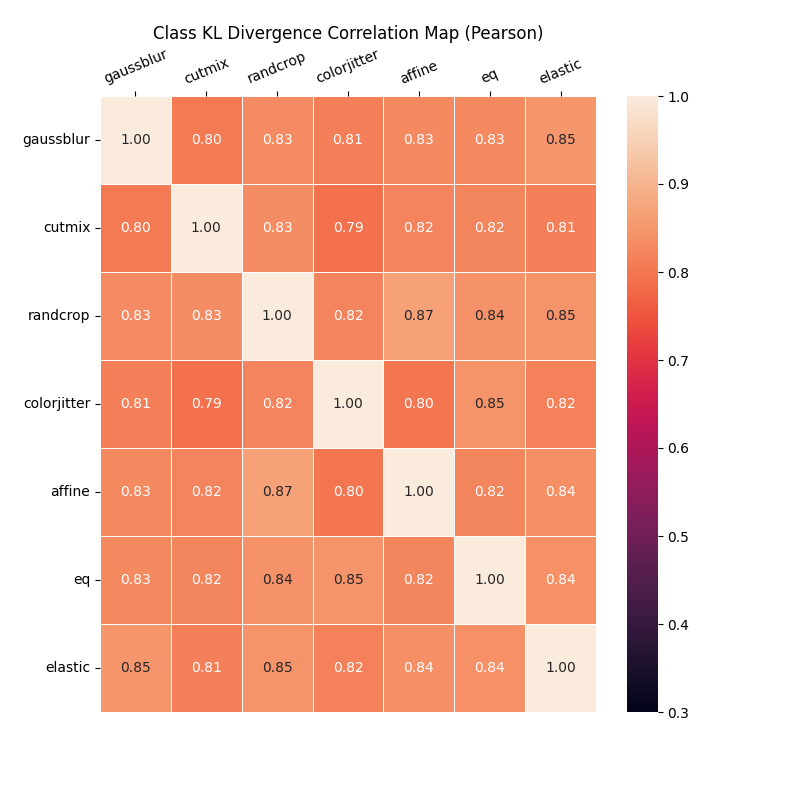}
\centering
\end{figure}

Lastly, we display the results for the Class-KLD metric. \autoref{fig:box:class_kld} shows the boxplot for the metric and \autoref{fig:cor:class_kld} shows its correlation map.

Class-KLD proved a hard metric to interpret. Due to the large amount of outliers, the visual density of the outliers, and most of the prediction metric values falling very close to one another, the boxplot (\autoref{fig:box:class_kld}) becomes convoluted and hard to analyze. An observation we can make from the boxplot is that most of the values for all augmentations tend to fall within the $0$ to $1$ range, which indicates that the class predictions made by the augmented models do not stray very far from those of the baseline models on average. 


We note that although augmentation methods alter the importance attributed by the last convolutional layer to the image pixels, the resulting models generate classification probability distributions that, in general, have a Class-KLD value in a relatively small range (even if the outliers are numerous). To examine this fact and the previous observation, it is important to remember that the final convolutional layer of our chosen architecture is followed by classification layers. This aspect of EfficientNet B0 suggests that these classification layers can approximate the resulting probabilities of the network in a relatively similar way between baseline and augmented models, even with the differences observed in the final convolutional layer through some of the CAM metrics (such as the overlap rate with a threshold of $Y=5$ as seen in \autoref{fig:box:overlap_5} and \autoref{fig:cor:overlap_5}). We also note that this behavior holds across all augmentations for Class-KLD due to the high correlation values between all augmentation pairs (\autoref{fig:cor:class_kld}).

\subsubsection{General Analysis}\label{ch:experiments:results:analysis}

As we can see from the metric distribution boxplots (\autoref{fig:box:overlap_20}, \autoref{fig:box:overlap_10}, \autoref{fig:box:overlap_5}, \autoref{fig:box:pearson}, \autoref{fig:box:spearman}, \autoref{fig:box:mae}, \autoref{fig:box:mse}, \autoref{fig:box:class_kld}), although they enable general analyses of the behavior of all augmentations, their individual results are similar across augmentations. This fact makes it challenging to discern meaningful differences between the behaviors of the augmentations by visual analysis of the boxplots alone; this is partly due to the boxplot distribution analysis not lending itself to image-by-image metric comparisons between the augmentations, but allowing only aggregate analyses. The augmentations behave very similarly to one another in these aggregate comparisons, and the correlation maps do not immediately provide much direct insight either. We also investigated segmenting the metric results for each augmentation, segmenting them by cases where both baseline and augmented models predicted the target label correctly, cases where both models classified the image incorrectly, and cases where one model predicted correctly while the other predicted an incorrect class. We did not identify any tendencies that could help us further differentiate behavior between augmentations through this method. A discussion about and some examples of these investigations can be found in \hyperref[ch:appendix:segmentation]{Appendix \ref*{ch:appendix:segmentation}}.

In the course of this work, we also conducted qualitative explorations into the CAMs across augmentations. The purpose was to gain a better idea of what the extremes for each metric would look like, but they ultimately did not contribute insights to the analyses we made in our experiment, even if they do provide a good visualization of what differences in CAMs can look like. Some of the results we obtained can be seen in  \hyperref[ch:appendix:cam]{Appendix \ref*{ch:appendix:cam}}.

Although hard to analyze, the differences in correlation across augmentations imply that augmentations present different behaviors. For example, the correlation for overlap rate with higher granularity (\autoref{fig:cor:overlap_5}) is significantly lower than for lower granularity (\autoref{fig:cor:overlap_20}). This observation indicates that this aspect of the relationship between augmentations follows less of a linear relationship than MAD (\autoref{fig:cor:mae}), MSD (\autoref{fig:cor:mse}), Pearson (\autoref{fig:cor:pearson}), and Spearman (\autoref{fig:cor:spearman}) correlation maps would indicate. In fact, as mentioned earlier, when focusing on general differences between CAMs, the correlation between augmented models and baseline is higher, but when focusing on the most significant regions for each augmentation, their correlation score is lower, which could indicate different patterns of behavior at scale.

We note that when analyzing the correlation maps in absolute terms, we observe that the correlation values are not significantly different across pairs. MAD's correlation map (\autoref{fig:cor:mae}), for example, has a minimum correlation between augmentations of 0.58 and a maximum of 0.67, which means the range of variation between correlation values is low. Additionally, in most metrics, the correlation maps score above 0.5, indicating moderate correlations between augmentations for those metrics.

Even though the absolute correlation values are not remarkably different, another analysis we can attempt is a relative one. By evaluating which pairs of augmentations are the most and least correlated across all metrics, supposing the existence of behavior profiles, then specific augmentation pairs should consistently score highly, while others would consistently score lower. To measure this, we count the number of times an augmentation pair is among the most strongly or most weakly correlated pair of augmentations across the correlation maps for each metric. For each CAM metric, if we count the four most and least correlated pairs for each metric's correlation map across augmentations and then aggregate the number of times those pairs appear across all metrics' correlation maps, we will have a relative analysis of the correlations. Following these steps, we get \autoref{tab:strongest} for the count of strongest pairs, and \autoref{tab:weakest} for the count of weakest pairs.

\begin{table}[h!]
\caption{Frequency table counting how many times each pair of augmentations appeared in the top four most strongly correlated augmentation pairs across all metrics' correlation maps}
\label{tab:strongest}
\begin{center}
\begin{tabular}{|l|l|l|l|l|}
\hline
Color Jitter - Affine Transform& 6 \\ \hline
Cutmix - Affine Transform	 & 6 \\ \hline
Cutmix - Color Jitter       & 5 \\ \hline
Gaussian Blur - Elastic Transform & 5 \\ \hline
Color Jitter - Elastic Transform & 3 \\ \hline
Gaussian Blur - Random Cropping & 2 \\ \hline
Color Jitter - Equalization & 1 \\ \hline 
Gaussian Blur - Elastic Transform & 1\\ \hline
Random Crop - Affine Transform & 1 \\ \hline 
Random Crop - Equalization	 & 1 \\ \hline 
Random Crop - Elastic Transform & 1 \\ \hline 


\end{tabular}
\end{center}
\end{table}

\begin{table}[h!]
\caption{Frequency table counting how many times each pair of augmentations appeared in the top four most weakly correlated augmentation pairs across all metrics' correlation maps}
\label{tab:weakest}
\begin{center}
\begin{tabular}{|l|l|l|l|l|}
\hline
Gaussian Blur - Cutmix	 & 8 \\ \hline
Cutmix - Equalization		 & 7 \\ \hline
Affine Transform - Equalization	 & 5 \\ \hline
Gaussian Blur - Affine Transform	& 3 \\ \hline 
Cutmix - Elastic Transform	 & 2 \\ \hline
Cutmix - Random Cropping	 & 2 \\ \hline 
Gaussian Blur - Equalization	 & 2 \\ \hline 
Cutmix - Color Jitter	 & 1 \\ \hline 
Cutmix - Affine Transform	 & 1 \\ \hline 
Color Jitter - Affine Transform	 & 1 \\ \hline 

\end{tabular}
\end{center}
\end{table}

Analyzing the table for strongest correlations (\autoref{tab:strongest}), we can tell that Cutmix-Affine appears in the top 4 most strongly correlated for 6 out of 8 metrics, Cutmix-Color Jitter appears 5 out of 8 times, and Color Jitter-Affine Transform appears 6 out of 8 times, with Gaussian Blur-Elastic Transform appearing 7 out of 8 times as well. These data points to Cutmix, Affine Transform, and Color Jitter having stronger correlations across the trio, possibly suggesting the existence of a sort of cluster of augmentation behaviors and that Gaussian Blur and Elastic Transform constitute another cluster or profile of impact. Analyzing the most weakly correlated pairs (\autoref{tab:weakest}), Gaussian Blur-Cutmix and Cutmix-Equalization are the least correlated pairs, with Gaussian Blur-Cutmix appearing 8 out of 8 times, Cutmix-Equalization appearing 7 out of 8, and Affine Transform-Equalization 5 out of 8 times. This observation supplements the earlier observation by suggesting that this Cutmix cluster and the Gaussian cluster have the weakest correlation among them and additionally implying that Equalization may be its own cluster due to it appearing in the four most weakly correlated pairs with both Gaussian and Cutmix 3 and 7 times out of 8, respectively. 

\section{Conclusion}\label{ch:conclusion}



In this work, our goal was to investigate a possible methodology to work with CAMs at scale for the purpose of data augmentation impact analysis. By proposing a generic series of steps and then applying them to a specific set of configurations, we have shown an initial exploration of how one may utilize this methodology, as well as the type of results it may be able to generate.


Metric distribution plots and correlation plots, for our specific selection of metrics, seem apt for generating data about the augmentations as a group. We find that for the overlap rate, the metric values get lower as the augmented models become more sensitive to different regions of the test images. The models also do not correlate strongly with each other through their overlap rate, and the correlation diminishes with an increase in region importance. Through MSD and MAD, we note that the average difference between augmented and baseline CAMs tends to follow the same distribution for all augmentations. They also present a moderate correlation (roughly in the range of $0.5$ to $0.6$) between augmentations and indicate that the augmentations are alike in the magnitude of differences between augmented and baseline CAMs. The Pearson and Spearman metrics show a median correlation at or below the moderate correlation zone for all augmentations, meaning more than half of the CAMs do not have evidence of significant correlation or similarity between augmented and baseline model CAMs, but neither do they have evidence of low or insignificant correlation. They have some correlation, but it is hard to extract information from this observation alone. Lastly, although Class-KLD metric distributions struggle to yield observations by themselves, when combined with the Class-KLD correlation maps they suggest that the dense classification layers of a model are able to approximate the image classification probability distribution in a similar way across augmentations and test images, even if the final convolutional layer contributes different importances for image regions between augmentations.

Although these techniques may be able to produce observations on the behavior of augmentations as groups, they struggle to generate observations about the behaviors of individual data augmentation methods. We note that there are indications of clusters of techniques with higher correlations with one another, but the absolute differences in the correlation values are small. It may be the case that the set of metrics we chose may not fully capture some significant aspect of the differences between CAM or that the data augmentation techniques themselves have a harder-to-measure impact on model behavior than other aspects of model training. Additional work could also be conducted with our experiment to analyze and gauge possible impacts of image bias in the model CAMs induced by the CIFAR-10 dataset, as we did not test the models against other datasets.

The methodology presented in this work provides an initial, generic, scalable series of steps that allow for quantitative exploration of behavioral aspects in the relationship between data augmentation techniques. Impacts on model behavior, in general, are not trivial to analyze. Although our work struggled to investigate the effects of data augmentation on an individual level, the steps delineated in this work constitute a method usable with other metrics, other augmentation methods, other datasets, and for purposes other than analyzing the impact of data augmentation. The purpose of the methodology is to provide a modifiable way to investigate model behavior through CAMs that also provides adaptability to examine other sources of impact different from data augmentation.

Future works could expand on this work and its methodology by focusing on different aspects of it. Utilizing other analysis techniques over the metrics, such as clustering analysis techniques, we may consider characteristics of the augmentation methods as features to investigate groupings of methods with similar characteristics. Another type of future work would be to use datasets that have some ground truth for important image regions, allowing the analysis of differences between CAMs and ground truth. Other works could explore different convolutional architectures and datasets, other techniques to determine the importance of pixels, and investigate the impact of applying multiple data augmentation techniques used in conjunction instead of individually, comparing CAMs between augmented models directly instead of through a baseline, or other analysis metrics to measure CAM similarities. These investigations can reveal different insights into the relationships between important image regions and different aspects of the learning process for CNNs.

\bibliographystyle{unsrtnat}
\bibliography{references}

\clearpage
\appendix
\raggedbottom
\section{CAM examples}\label{ch:appendix:cam}
In this section, we share some of the CAMs we obtained after training. To generate these qualitative analyses of CAM images, we analyzed the metric values of all augmentations to find the test images whose CAM metrics had the highest mean or standard deviation across augmentations. These grid comparisons between CAMs help us visualize the extremes for each metric, as well as have a better idea of what CAM differences might visually look like.

\begin{figure}[H]
\caption{Grid comparing the model CAMs for the test image that produced the highest mean for the overlap rate (20) metric across all test images and augmentations.}
\label{fig:cam:overlap_rate_20_mean}
\includegraphics[width=.95\linewidth]{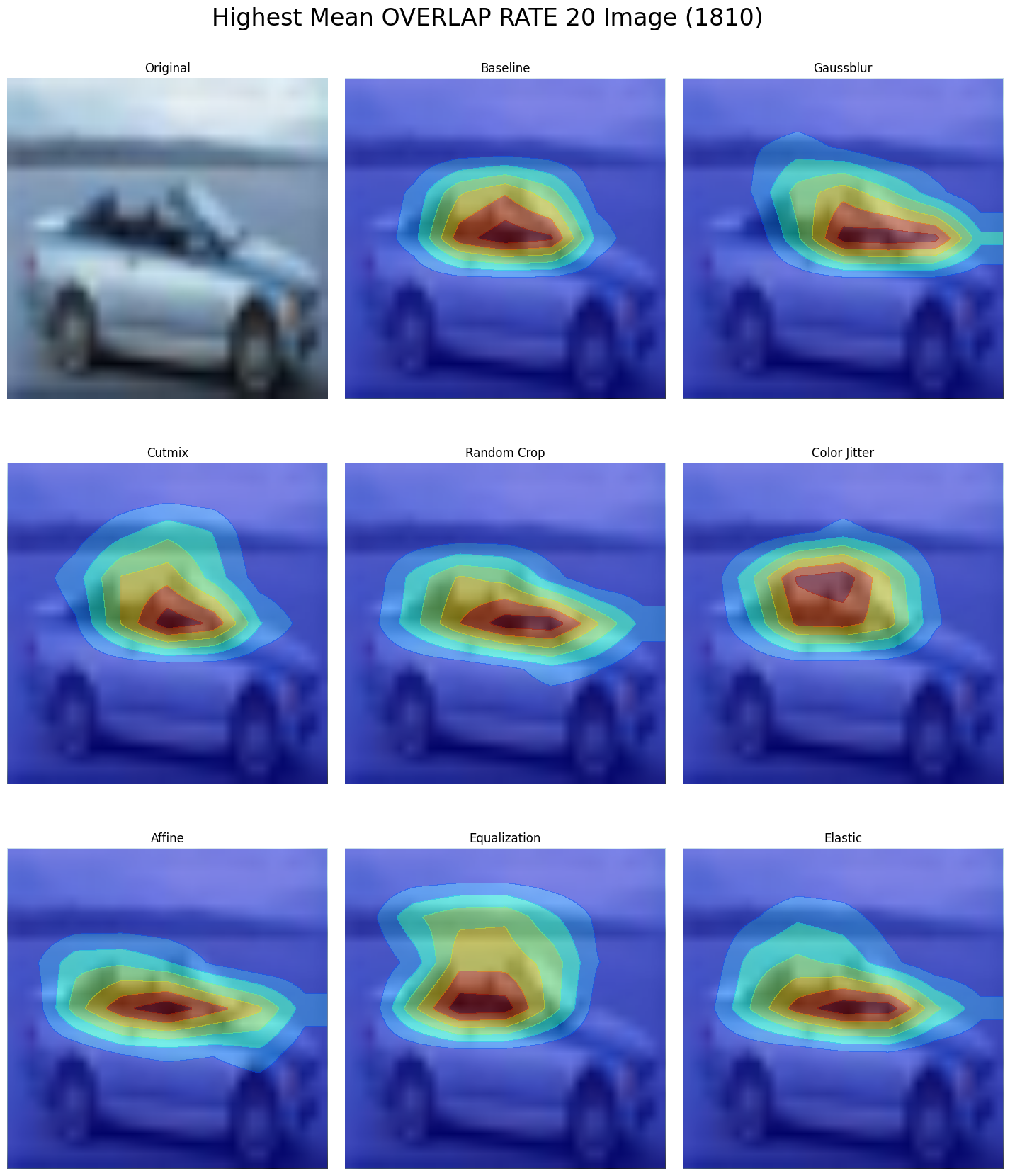}
\centering
\end{figure}

\begin{figure}[H]
\caption{Grid comparing the model CAMs for the test image that produced the highest standard deviation for the overlap rate (20) metric across all test images and augmentations.}
\label{fig:cam:overlap_rate_20_stdev}
\includegraphics[width=.95\linewidth]{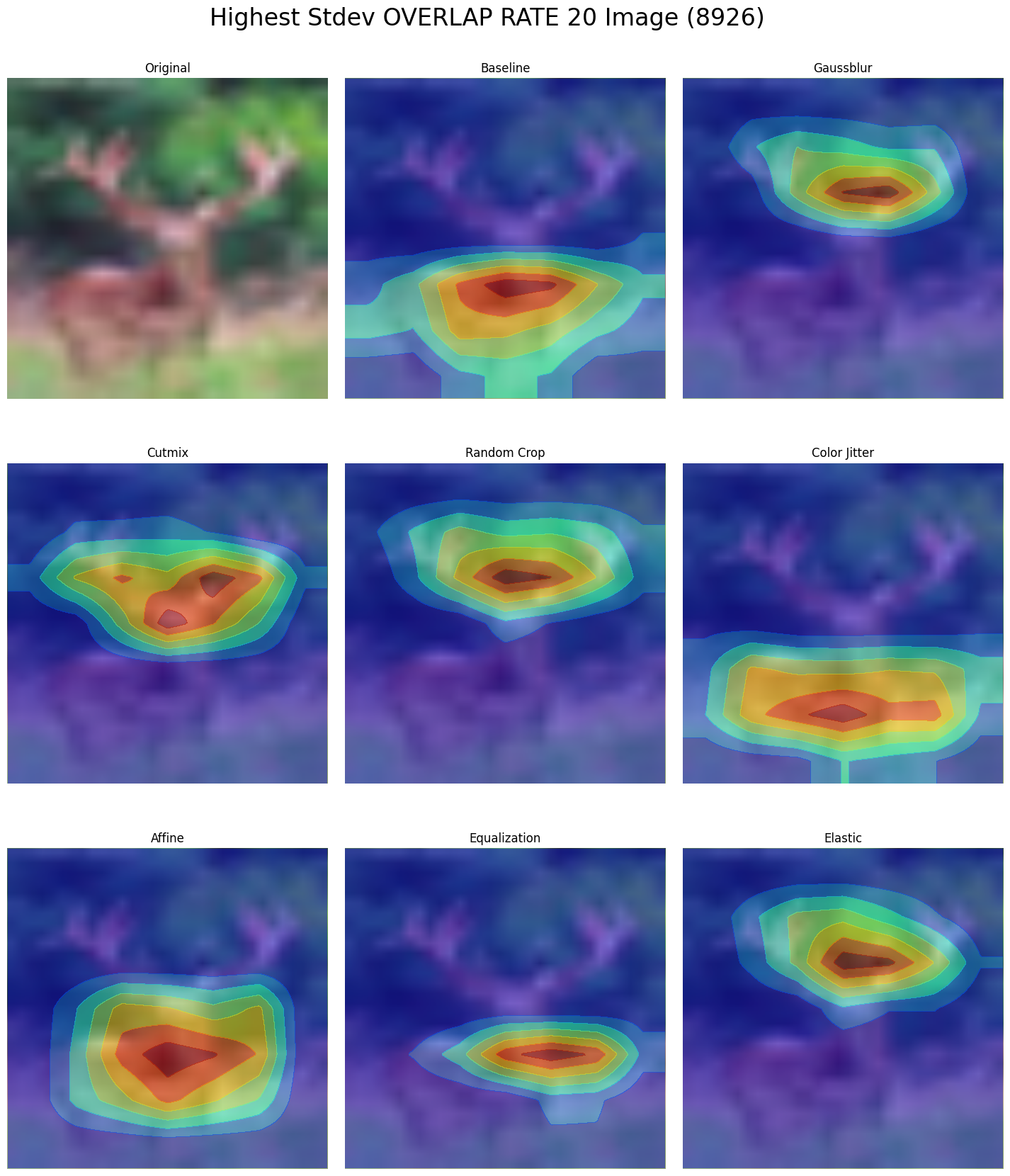}
\centering
\end{figure}

\begin{figure}[H]
\caption{Grid comparing the model CAMs for the test image that produced the highest mean for the overlap rate (10) metric across all test images and augmentations.}
\label{fig:cam:overlap_rate_10_mean}
\includegraphics[width=.95\linewidth]{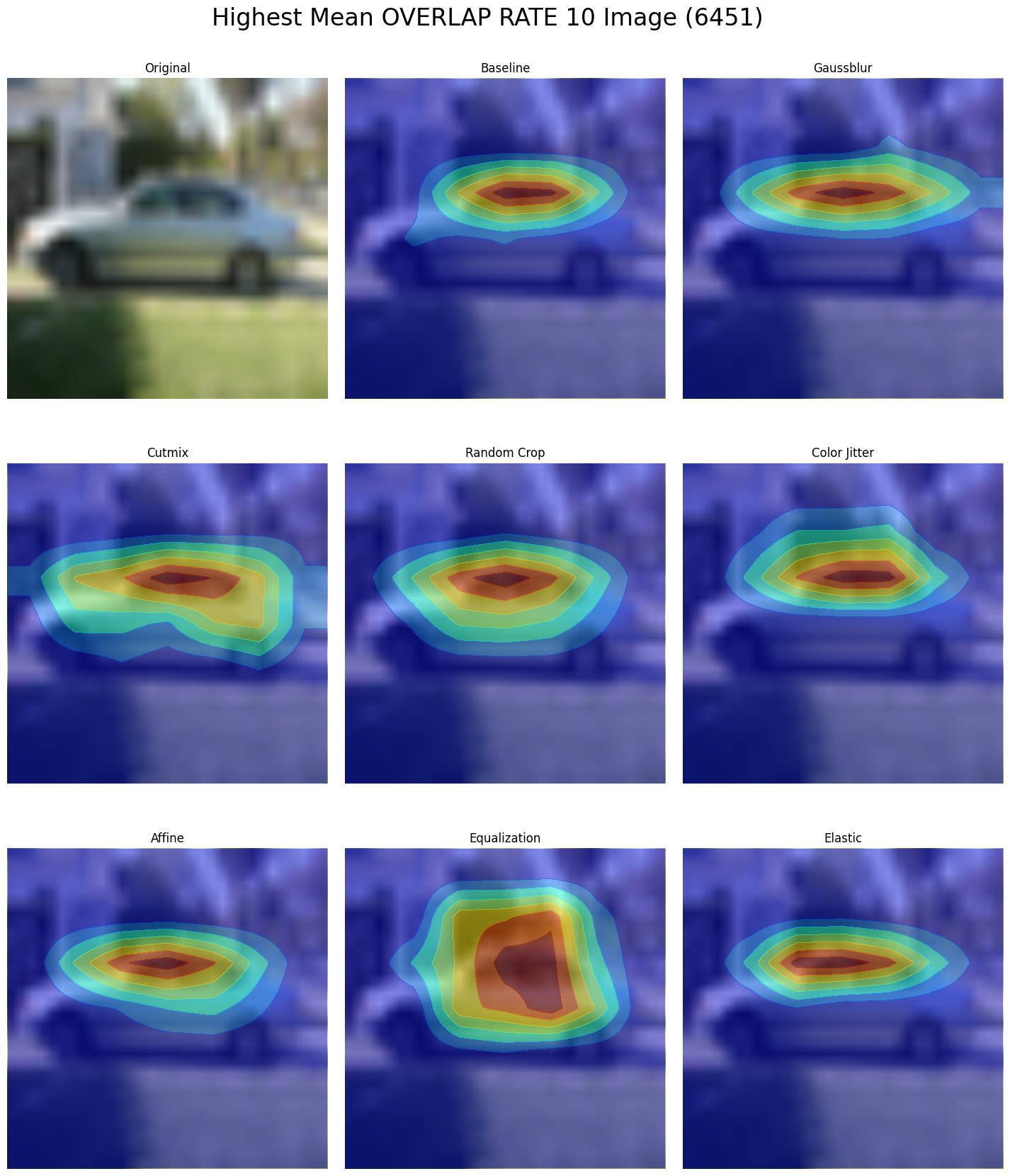}
\centering
\end{figure}

\begin{figure}[H]
\caption{Grid comparing the model CAMs for the test image that produced the highest standard deviation for the overlap rate (10) metric across all test images and augmentations.}
\label{fig:cam:overlap_rate_10_stdev}
\includegraphics[width=.95\linewidth]{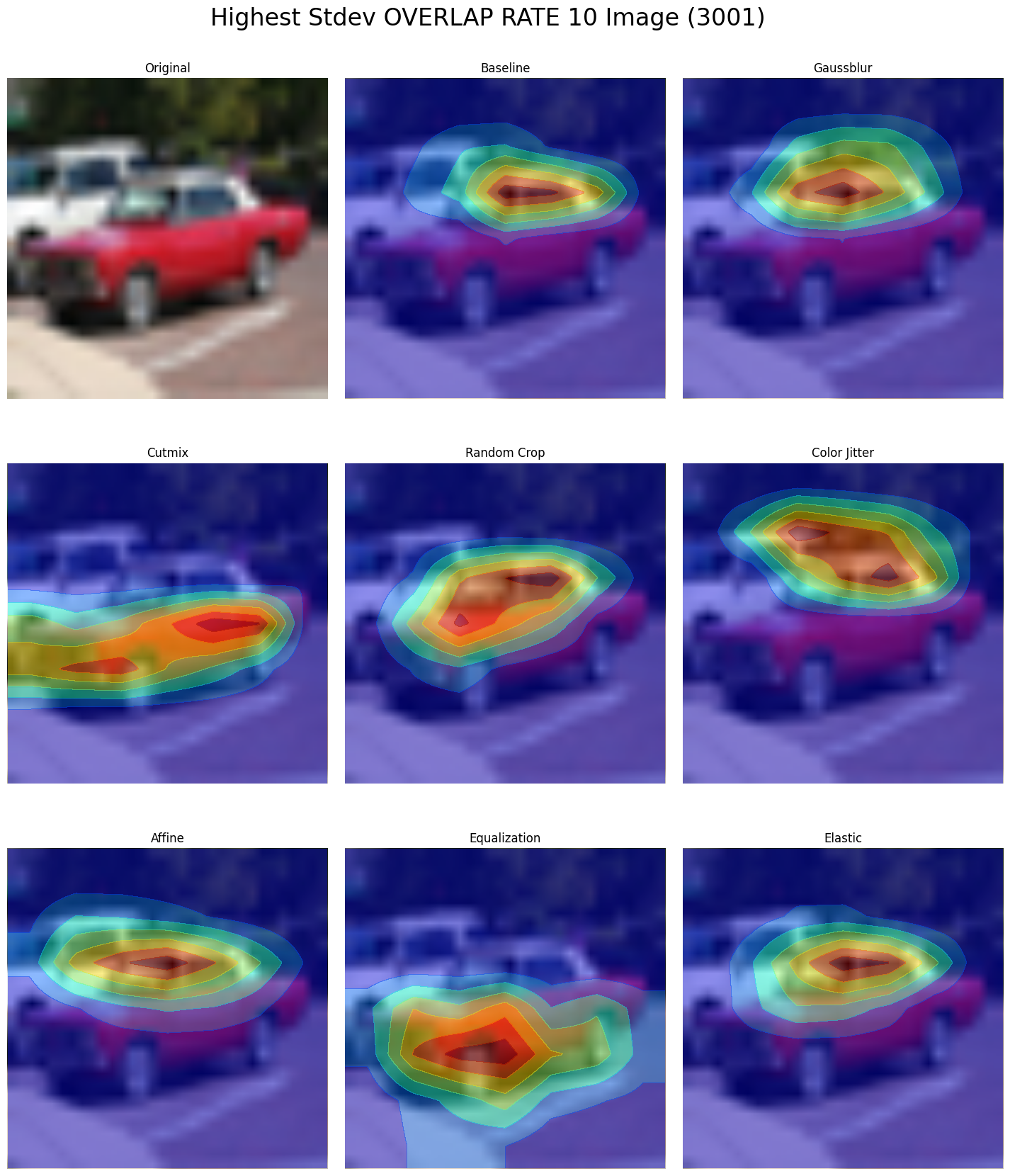}
\centering
\end{figure}

\begin{figure}[H]
\caption{Grid comparing the model CAMs for the test image that produced the highest mean for the overlap rate (5) metric across all test images and augmentations.}
\label{fig:cam:overlap_rate_5_mean}
\includegraphics[width=.95\linewidth]{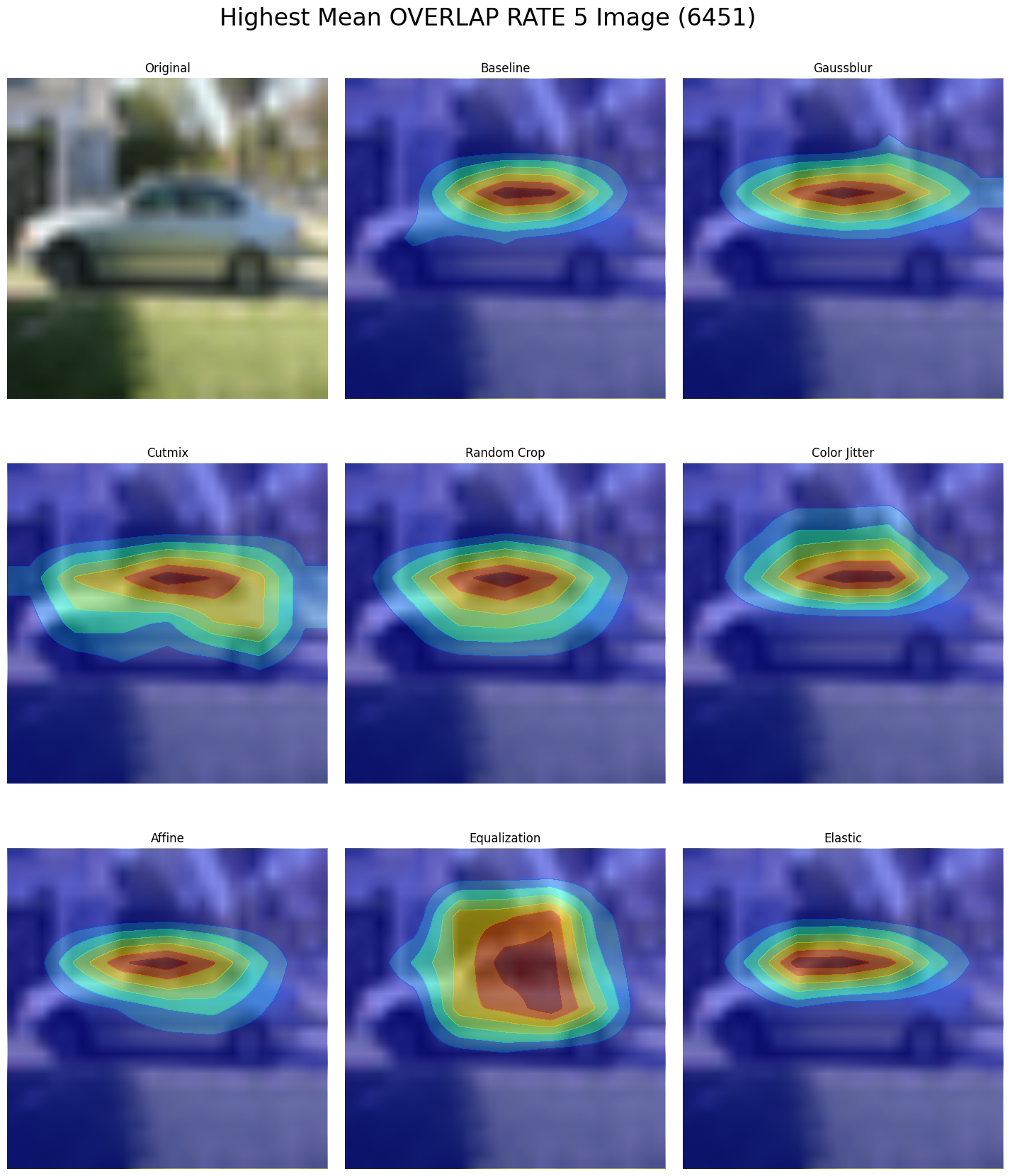}
\centering
\end{figure}

\begin{figure}[H]
\caption{Grid comparing the model CAMs for the test image that produced the highest standard deviation for the overlap rate (5) metric across all test images and augmentations.}
\label{fig:cam:overlap_rate_5_stdev}
\includegraphics[width=.95\linewidth]{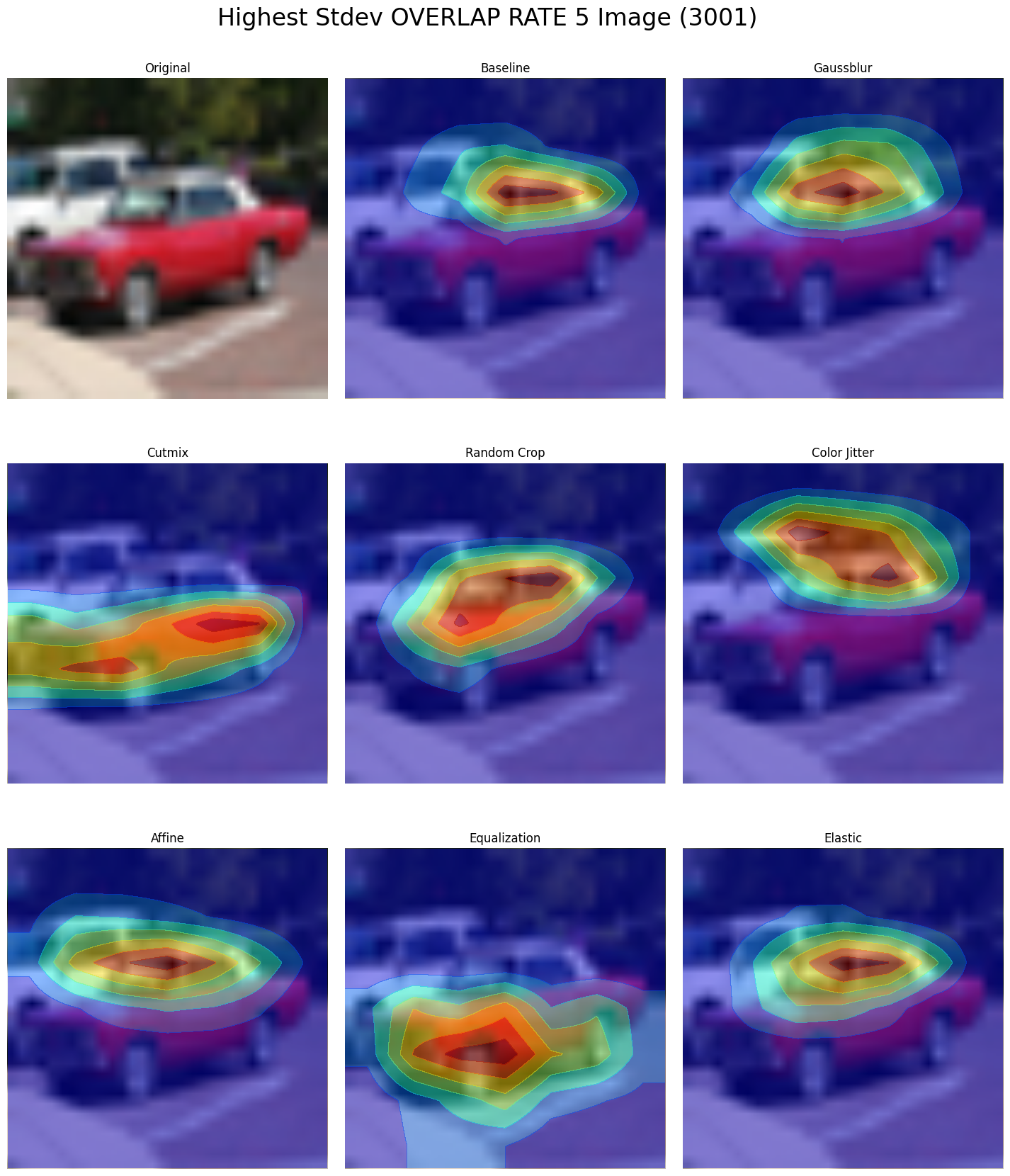}
\centering
\end{figure}

\begin{figure}[H]
\caption{Grid comparing the model CAMs for the test image that produced the highest mean for the Pearson Correlation metric across all test images and augmentations.}
\label{fig:cam:pearson_mean}
\includegraphics[width=.95\linewidth]{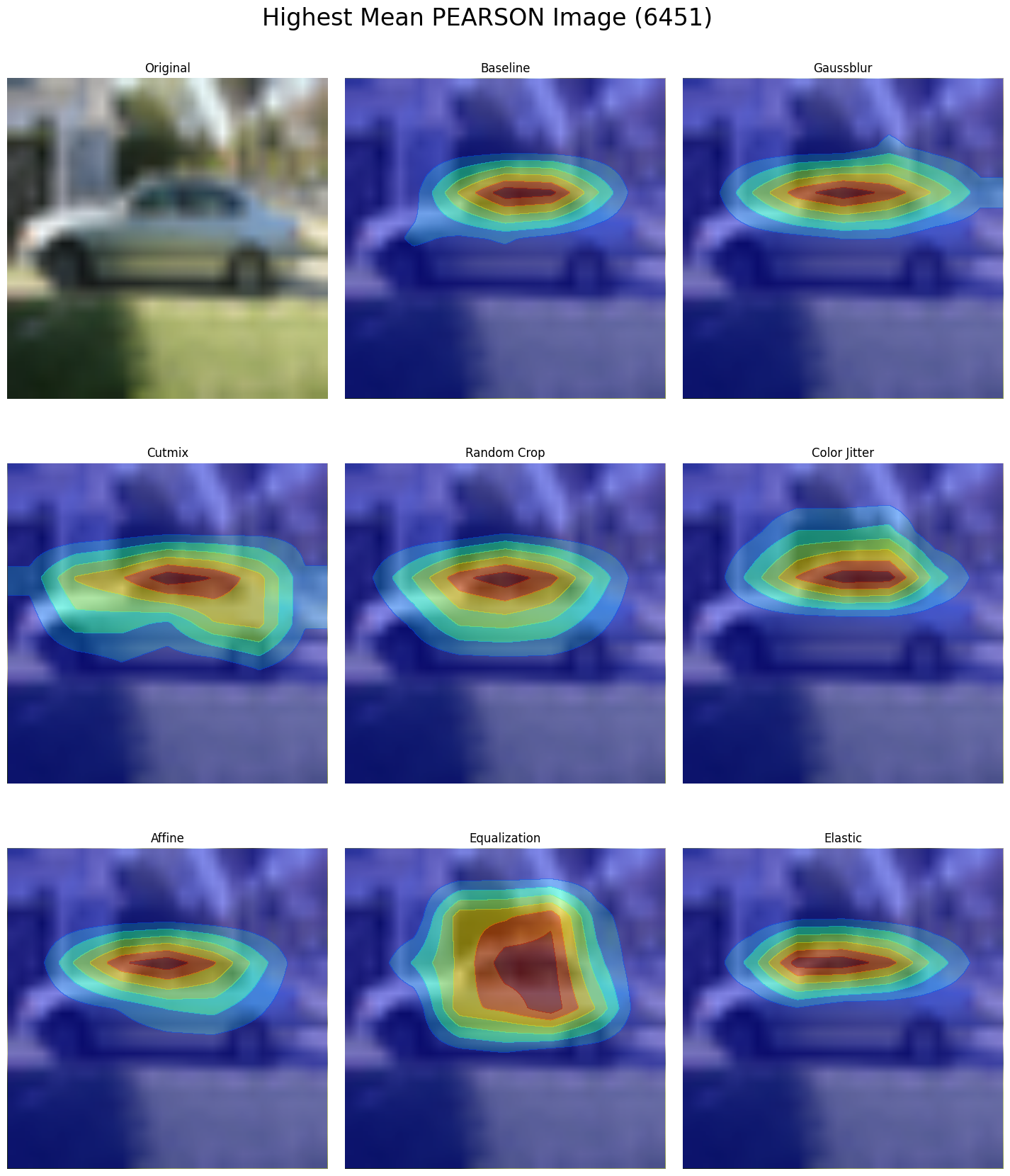}
\centering
\end{figure}

\begin{figure}[H]
\caption{Grid comparing the model CAMs for the test image that produced the highest standard deviation for the Pearson Correlation metric across all test images and augmentations.}
\label{fig:cam:pearson_stdev}
\includegraphics[width=.95\linewidth]{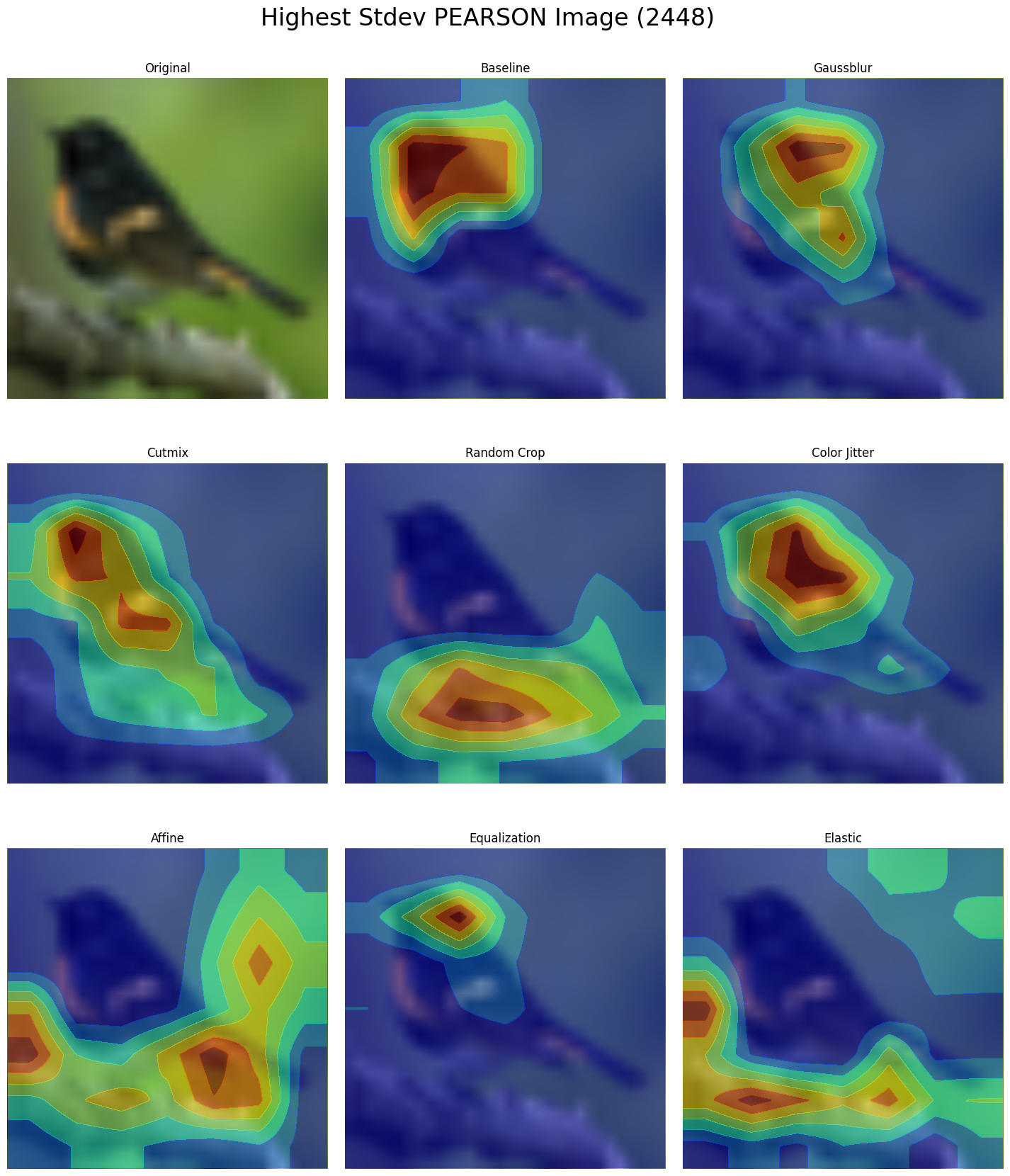}
\centering
\end{figure}

\begin{figure}[H]
\caption{Grid comparing the model CAMs for the test image that produced the highest mean for the Spearman Correlation metric across all test images and augmentations.}
\label{fig:cam:spearman_mean}
\includegraphics[width=.95\linewidth]{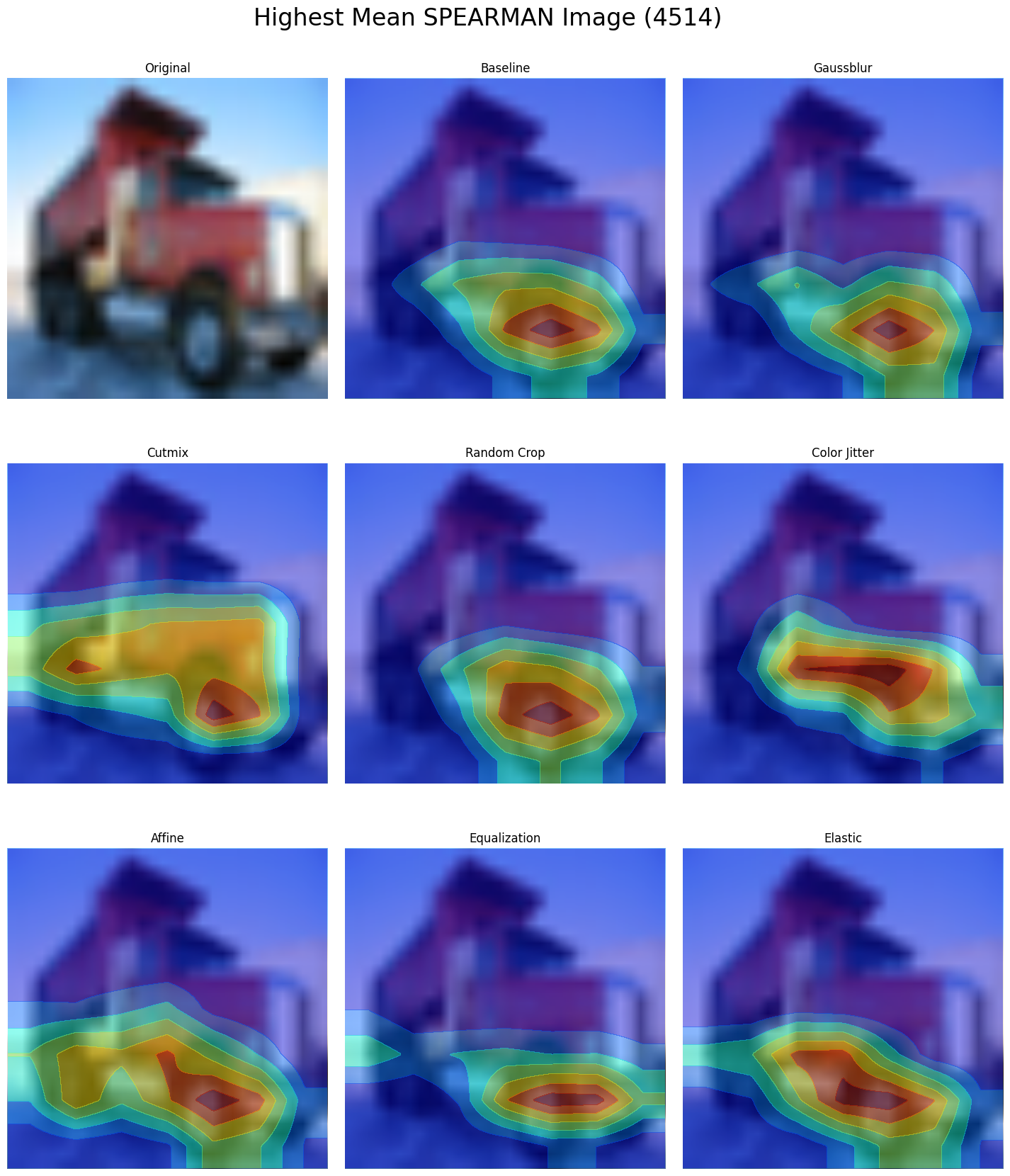}
\centering
\end{figure}

\begin{figure}[H]
\caption{Grid comparing the model CAMs for the test image that produced the highest standard deviation for the Spearman Correlation metric across all test images and augmentations.}
\label{fig:cam:spearman_stdev}
\includegraphics[width=.95\linewidth]{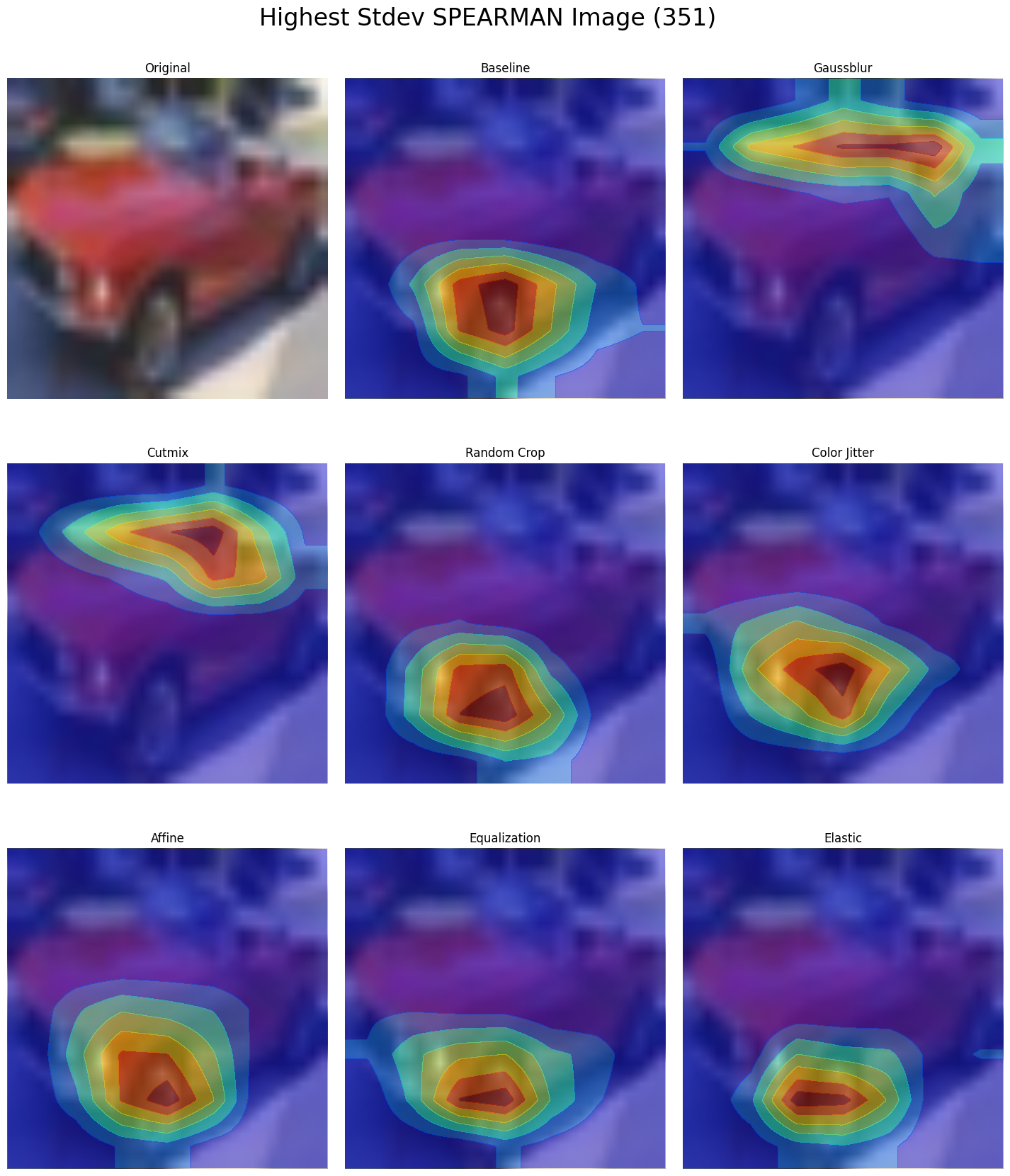}
\centering
\end{figure}

\begin{figure}[H]
\caption{Grid comparing the model CAMs for the test image that produced the highest mean for the MAD metric across all test images and augmentations.}
\label{fig:cam:mad_mean}
\includegraphics[width=.95\linewidth]{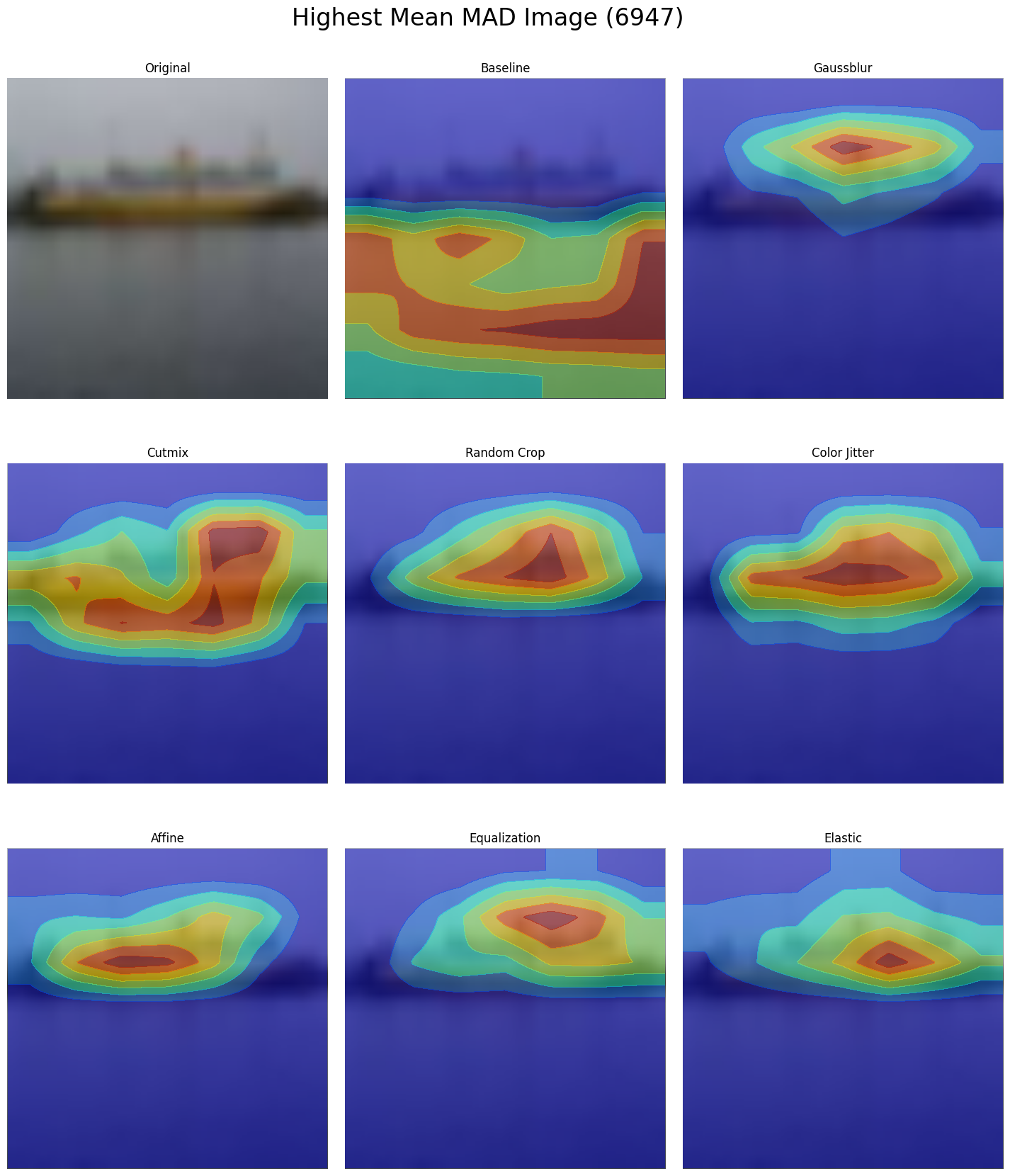}
\centering
\end{figure}

\begin{figure}[H]
\caption{Grid comparing the model CAMs for the test image that produced the highest standard deviation for the MAD metric across all test images and augmentations.}
\label{fig:cam:mad_stdev}
\includegraphics[width=.95\linewidth]{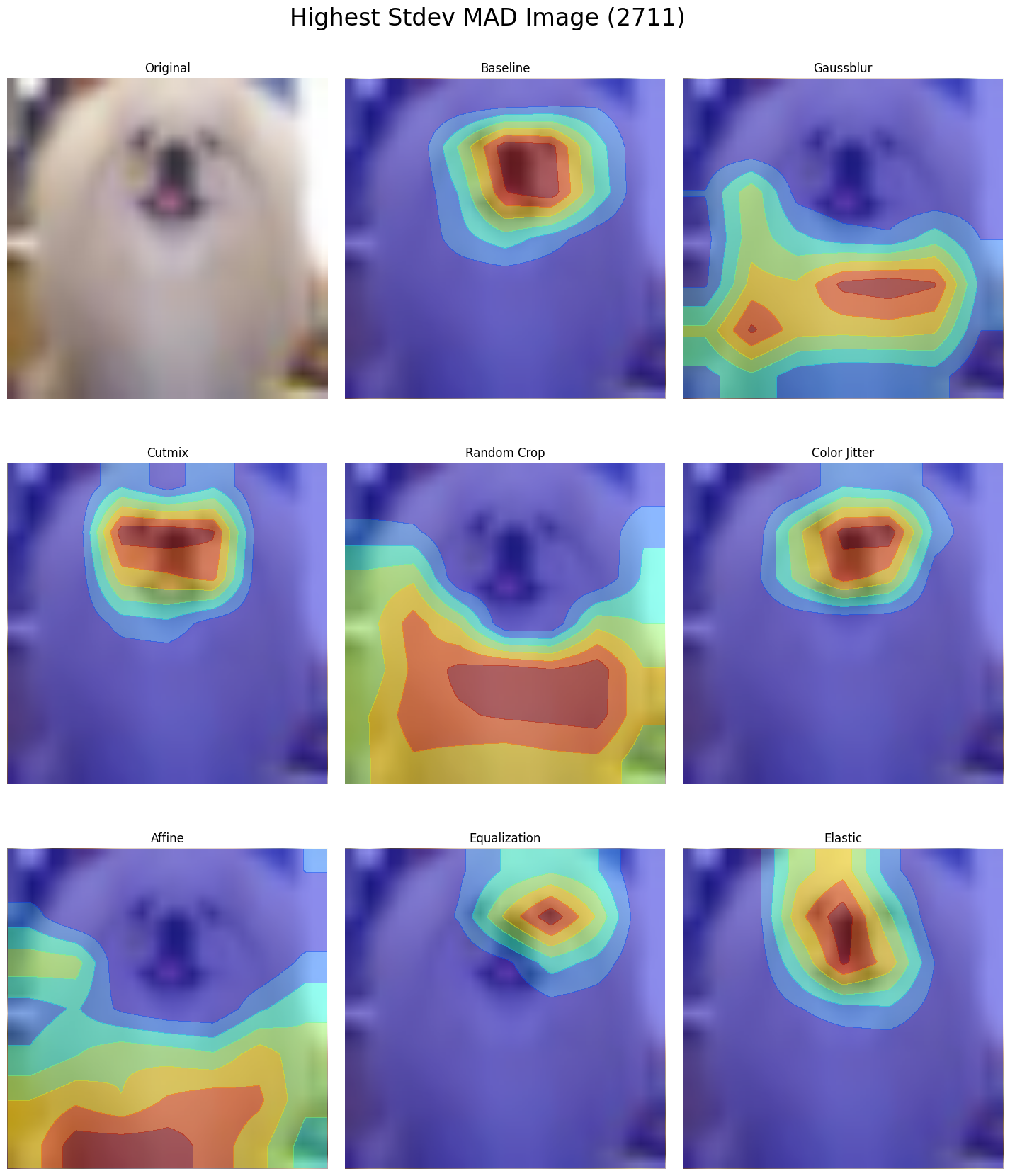}
\centering
\end{figure}

\begin{figure}[H]
\caption{Grid comparing the model CAMs for the test image that produced the highest mean for the MSD metric across all test images and augmentations.}
\label{fig:cam:msd_mean}
\includegraphics[width=.95\linewidth]{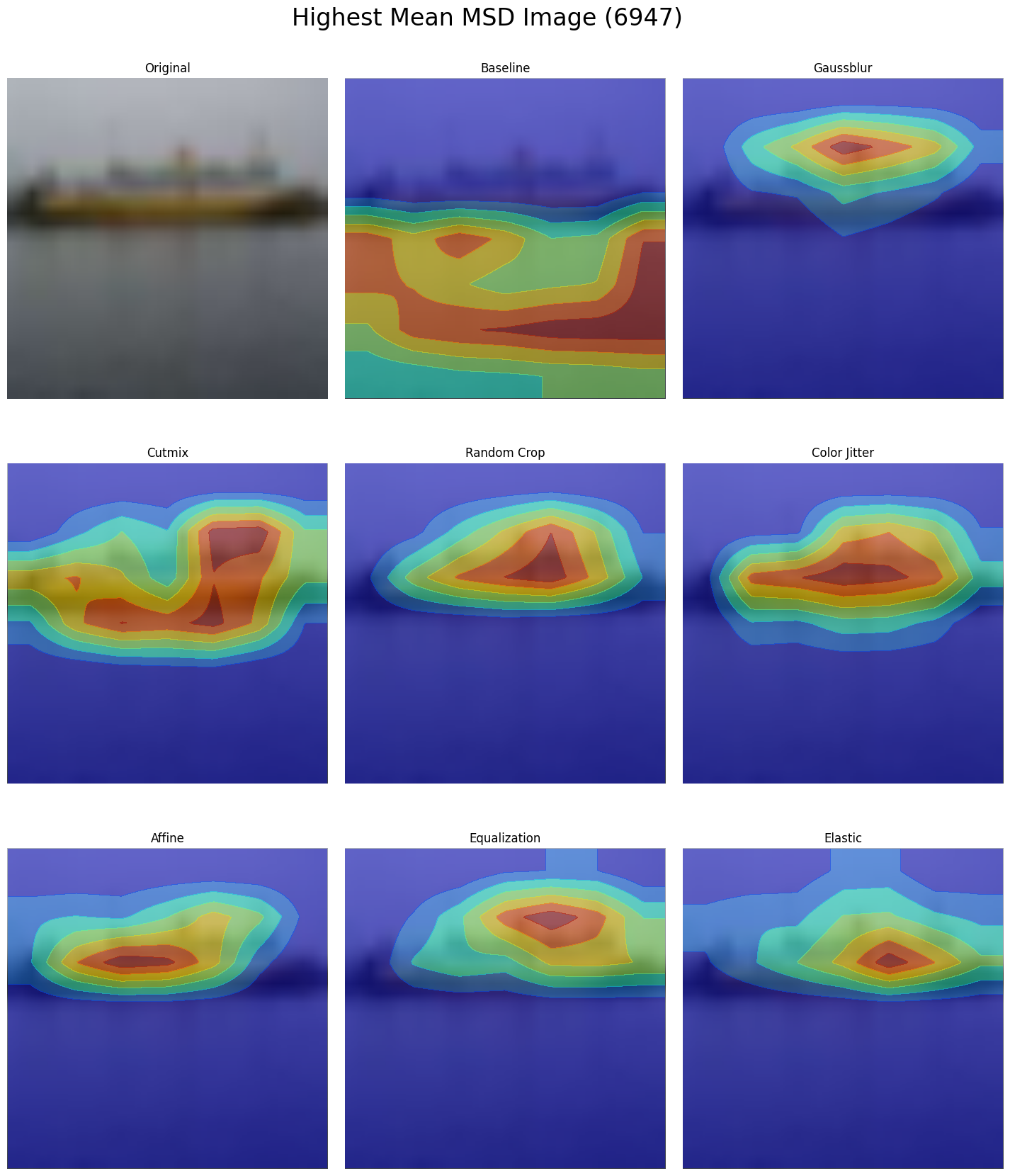}
\centering
\end{figure}

\begin{figure}[H]
\caption{Grid comparing the model CAMs for the test image that produced the highest standard deviation for the MSD metric across all test images and augmentations.}
\label{fig:cam:msd_stdev}
\includegraphics[width=.95\linewidth]{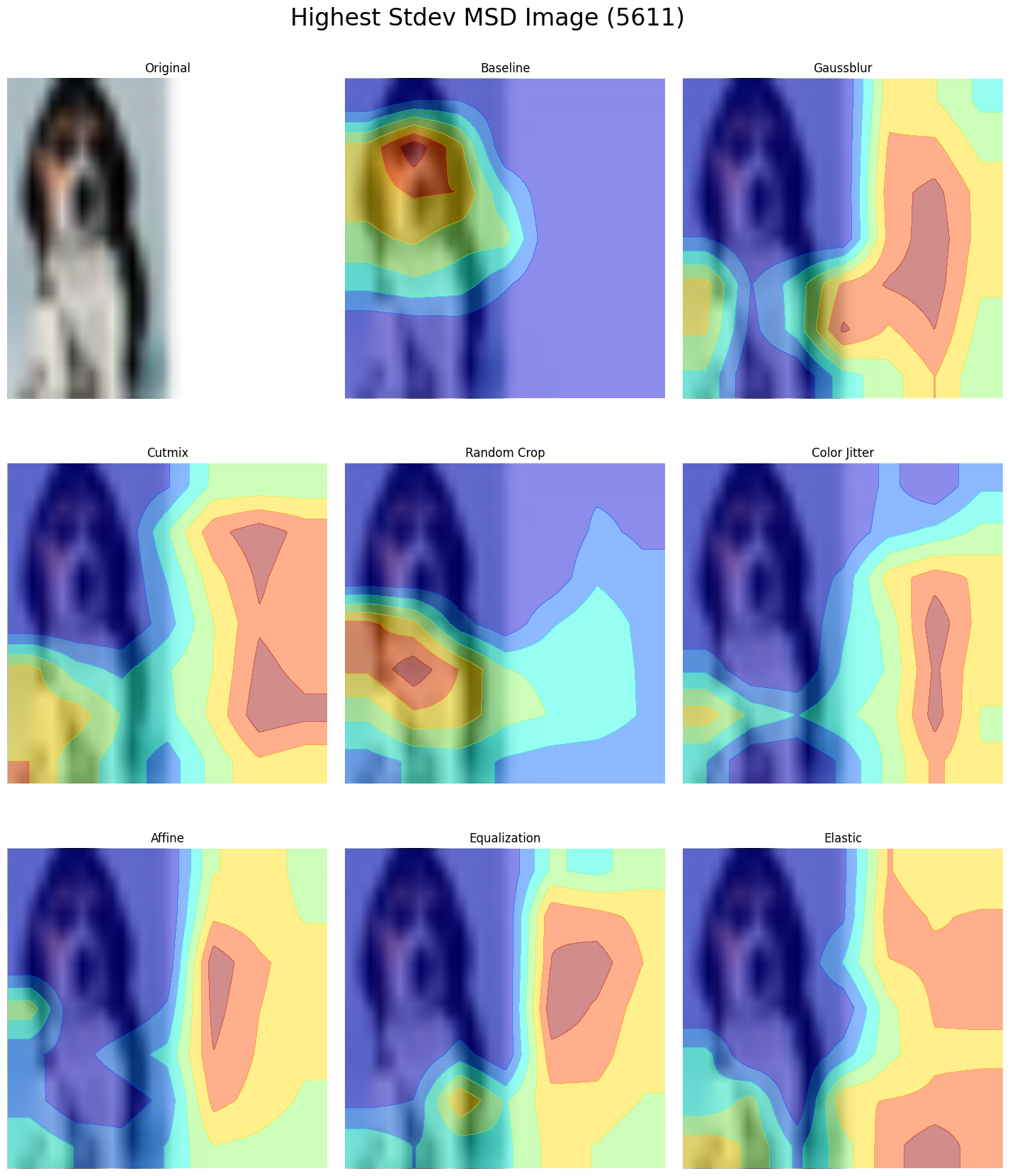}
\centering
\end{figure}

\begin{figure}[H]
\caption{Grid comparing the model CAMs for the test image that produced the highest mean for the Class-KLD metric across all test images and augmentations.}
\label{fig:cam:class_kld_mean}
\includegraphics[width=.9\linewidth]{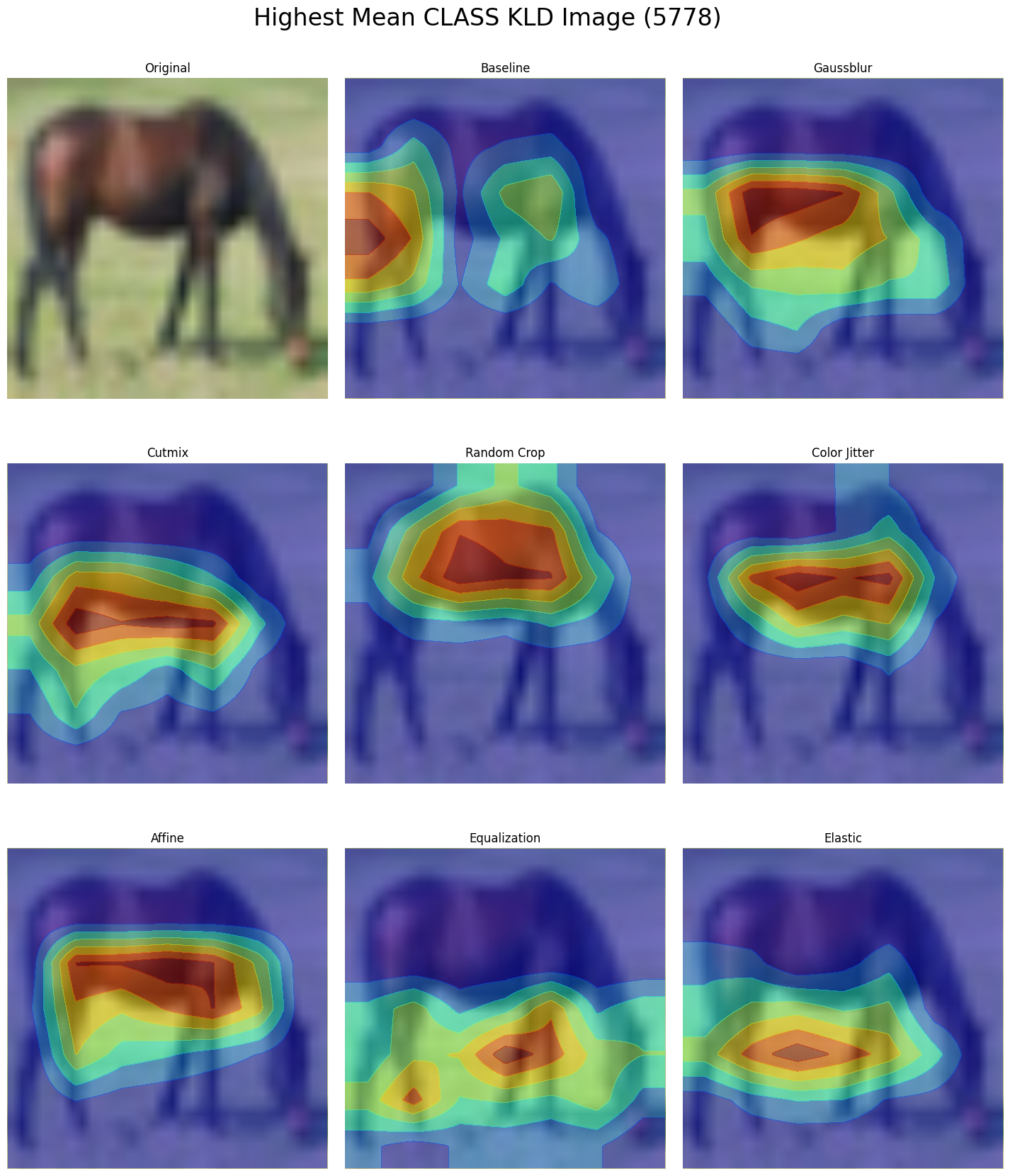}
\centering
\end{figure}

\begin{figure}[H]
\caption{Grid comparing the model CAMs for the test image that produced the highest standard deviation for the Class-KLD metric across all test images and augmentations.}
\label{fig:cam:class_kld_stdev}
\includegraphics[width=.9\linewidth]{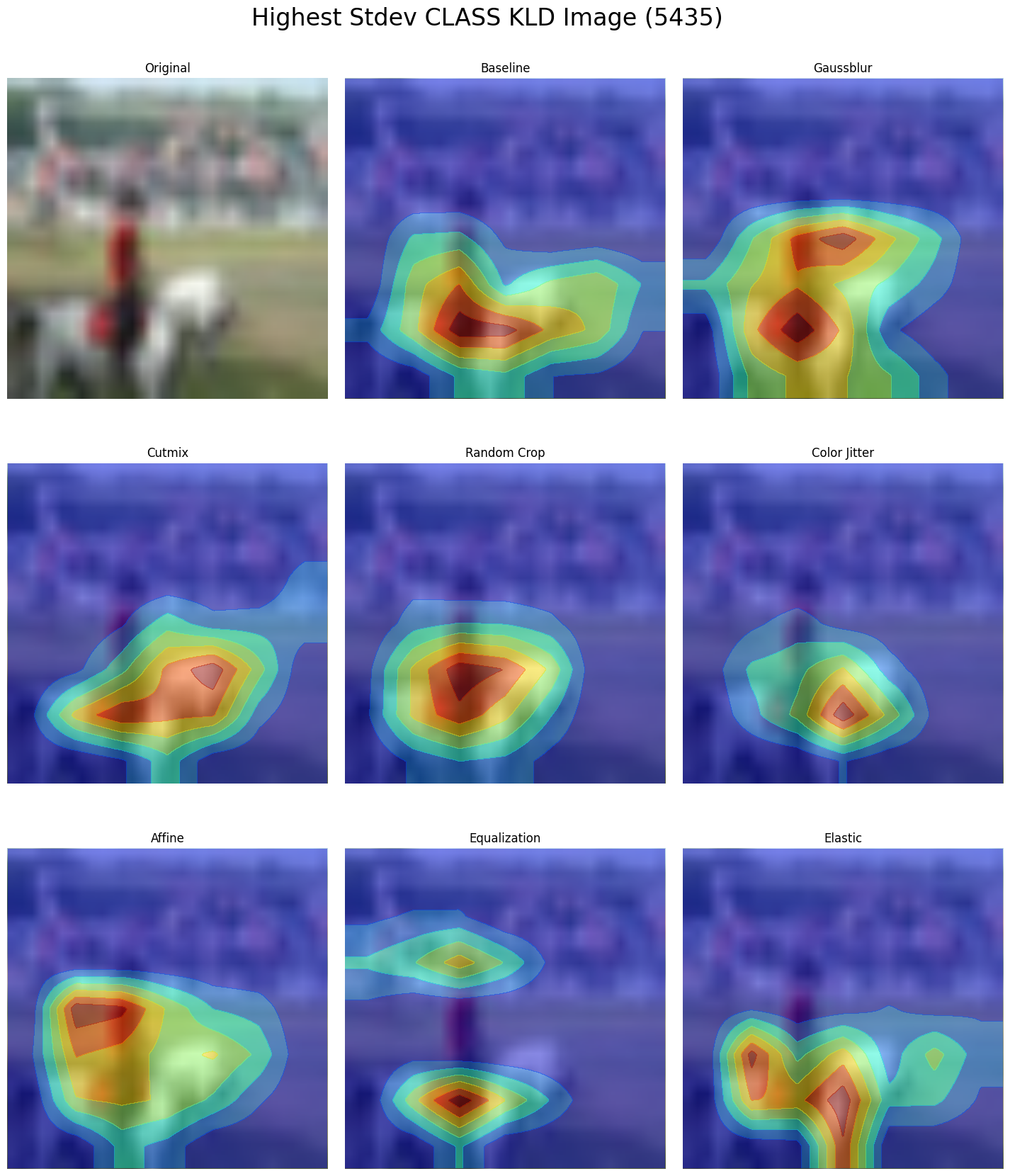}
\centering
\end{figure}

\autoref{fig:cam:overlap_rate_20_mean} shows the CAM comparison for the image with the highest mean for overlap rate (20), \autoref{fig:cam:overlap_rate_20_stdev} for highest standard deviation for overlap rate (20), \autoref{fig:cam:overlap_rate_10_mean} for highest mean for overlap rate (10), \autoref{fig:cam:overlap_rate_10_stdev} for the highest standard deviation for overlap rate (10), \autoref{fig:cam:overlap_rate_5_mean} for highest mean for overlap rate (5), \autoref{fig:cam:overlap_rate_5_stdev} for the highest standard deviation for overlap rate (5), \autoref{fig:cam:pearson_mean} shows this comparison for the image with highest mean Pearson Correlation, \autoref{fig:cam:pearson_stdev} does so for the highest standard deviation for Pearson Correlation, \autoref{fig:cam:spearman_mean} for highest mean Spearman Correlation, \autoref{fig:cam:spearman_stdev} for highest standard deviation for Spearman Correlation, \autoref{fig:cam:mad_mean} shows the same comparison for the image with the highest mean MAD, \autoref{fig:cam:mad_stdev} does so for the highest standard deviation for MAD, \autoref{fig:cam:msd_mean} shows the comparison for highest mean MSD, \autoref{fig:cam:msd_stdev} does so for the highest standard deviation for MSD, \autoref{fig:cam:class_kld_mean} shows the comparison for highest mean Class-KLD, and
\autoref{fig:cam:class_kld_stdev} shows the CAM comparison for the image with the highest standard deviation for the Class-KLD metric.

\section{Controlling undesired sources of impact}\label{ch:appendix:determinism}

As we are interested solely in analyzing the effects of data augmentation, part of our work consisted of controlling for undesired sources of impact during the construction of the augmented datasets and the training of our models, such that the impact differences we observed would come only from the data augmentation methods. 

To accomplish this, we first needed to ensure that the algorithms from the neural network libraries we used behaved deterministically. To do so, we explicitly turned off non-deterministic behavior in both PyTorch and CUDA libraries, but to guarantee that other random generation functionality present in the augmentations behaved deterministically, we also manually seeded Numpy, Random, Torch, and Torchvision libraries wherever needed, including during the instantiation of the initial weights for the models. Additionally, when using Torch dataloaders to load batches of data from the datasets, the shuffle functionality was turned off, guaranteeing that models would load batches of images in the order they appear in the respective dataset.

After ensuring that these impact sources were controlled for, the next element that needed looking into was the construction of the augmented datasets and the ordering of the images therein. Although it would be impossible to control the image content of the datasets, as each augmentation will generate its own set of augmented images, it was important to control the order in which the images were ordered within the dataset.

As mentioned previously, every model uses the same train-validation split based on the training part of the CIFAR10 dataset. Of the 50,000 training images, 45,000 are used for training and 5,000 for validation of each epoch. Once we establish the baseline train dataset, an initial, purely augmented dataset is instantiated from the baseline dataset by generating one augmented image for every baseline image. 

This process generates images in the order they appear in the baseline dataset, such that the 45,000 augmented images follow the same indexing order as the baseline, making it so that an image with index 42 in the augmented dataset will use the image with index 42 of the baseline dataset as the source for its augmentation. 

For each augmentation method, we then concatenate the augmented images onto a copy of the baseline dataset for a total of 90,000 training images in this augmented dataset. After this, we shuffle the 90,000 images according to a specific, separate, stable seed. The seed used for this sampling is the same for every augmentation method, ensuring that the image index order will be the exact same across all augmented datasets. Note that the sampling seed is separate from the starting state seeds.

\section{Segmented results}\label{ch:appendix:segmentation}

Segmentation of the metric results was motivated by an issue we predicted could arise from the choice of using the predicted label for Grad-CAM instead of the ground truth label. Given that it would be reasonable to suppose each class has its own potentially distinct identifying features, if two different models predicted different labels for a given image, then it would be reasonable to expect their CAMs to be more dissimilar than they would be if they'd both predicted the same class. This case would be worrisome insofar as it could add hard-to-interpret data points into the CAM metric distributions. If, for example, two models predicted a large number of test images incorrectly, and both models made different predictions from one another at every step, then we could expect the metric results to be significantly different from each other, and it could become hard to analyze the differences in their patterns. 

To measure the impact of these cases, we ran experiments segmenting our results into combinations of both models being correct, one model predicting correctly and the other being wrong, as well as both being wrong, and found the impact of these cases on the metric distributions to be minimal for all metrics. As the boxplots for these results are numerous and do not show different patterns across metrics, we show a selection of these segmented boxplots for one specific metric and all augmentations.

\begin{figure}[H]
\caption{Segmented boxplot for Affine Transform and Overlap Rate (20), showing the distribution of the metric values across segmentations.}
\label{fig:seg:affine}
\includegraphics[width=.75\linewidth]{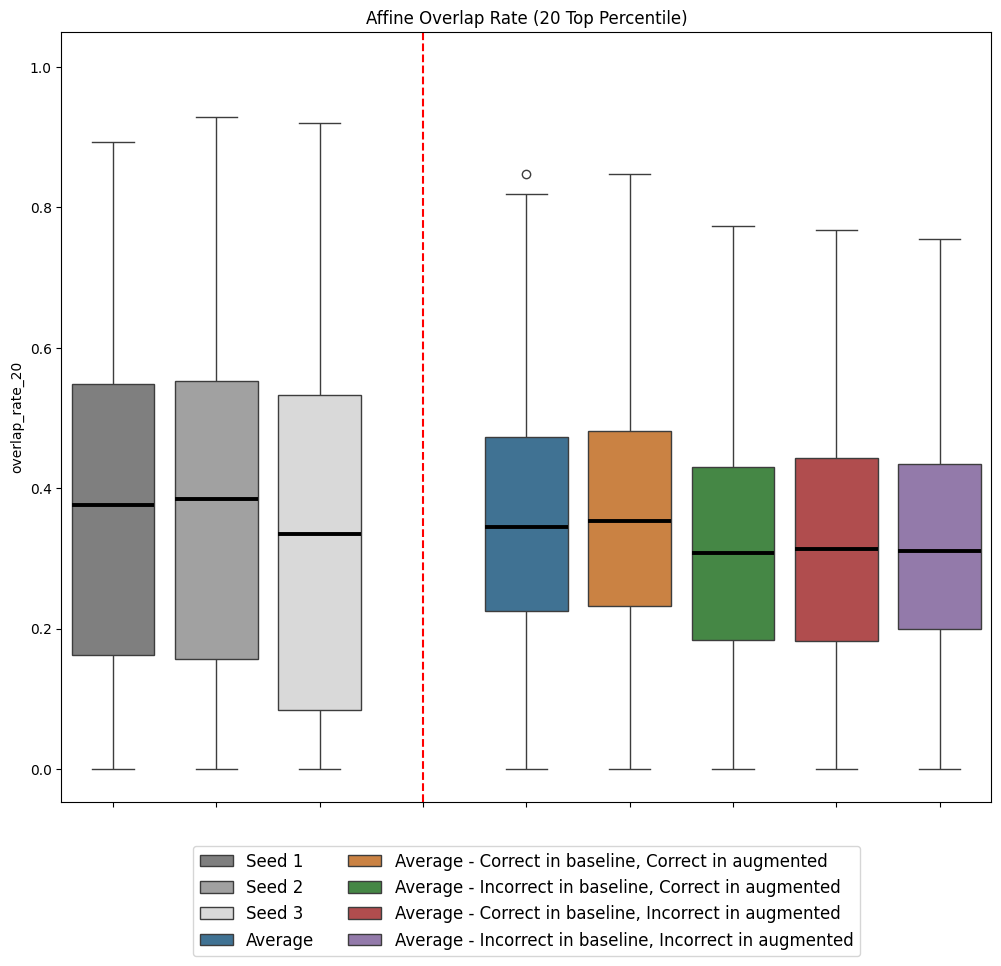}
\centering
\end{figure}

\begin{figure}[H]
\caption{Segmented boxplot for Color Jitter and Overlap Rate (20), showing the distribution of the metric values across segmentations.}
\label{fig:seg:colorjitter}
\includegraphics[width=.75\linewidth]{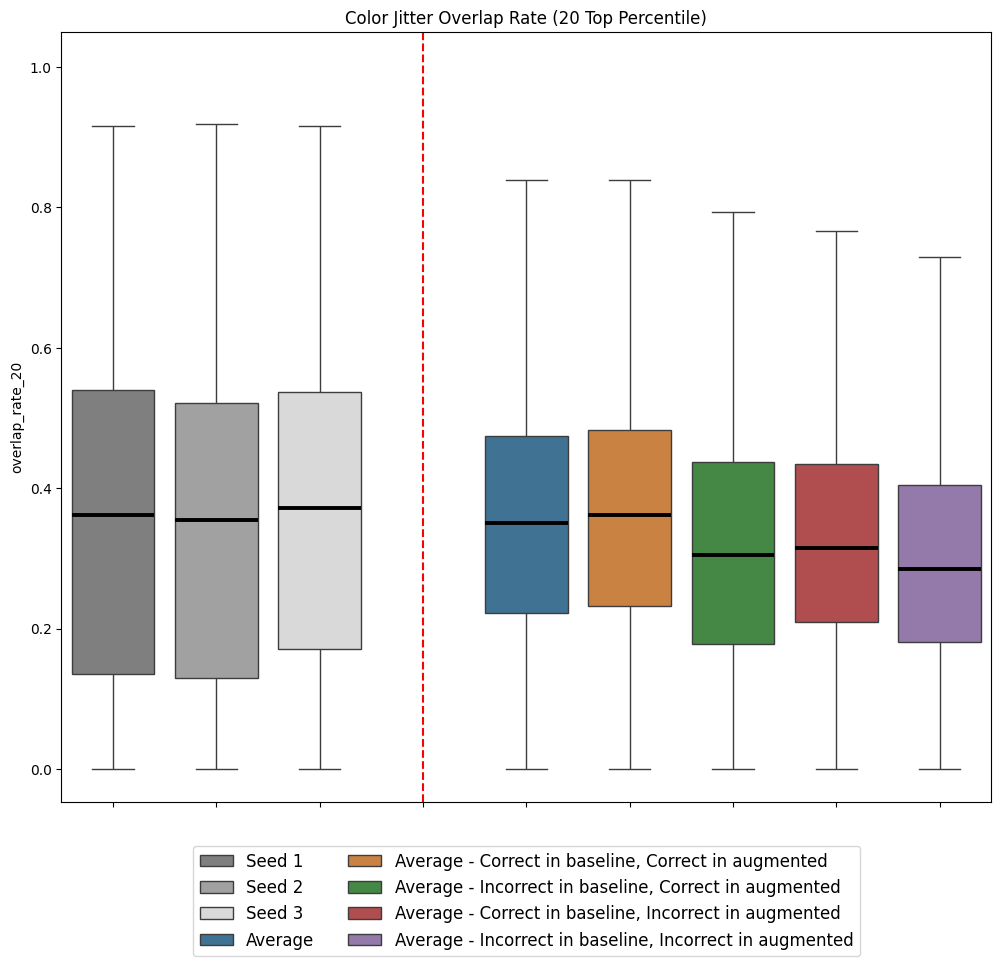}
\centering
\end{figure}

\begin{figure}[H]
\caption{Segmented boxplot for Cutmix and Overlap Rate (20), showing the distribution of the metric values across segmentations.}
\label{fig:seg:cutmix}
\includegraphics[width=.75\linewidth]{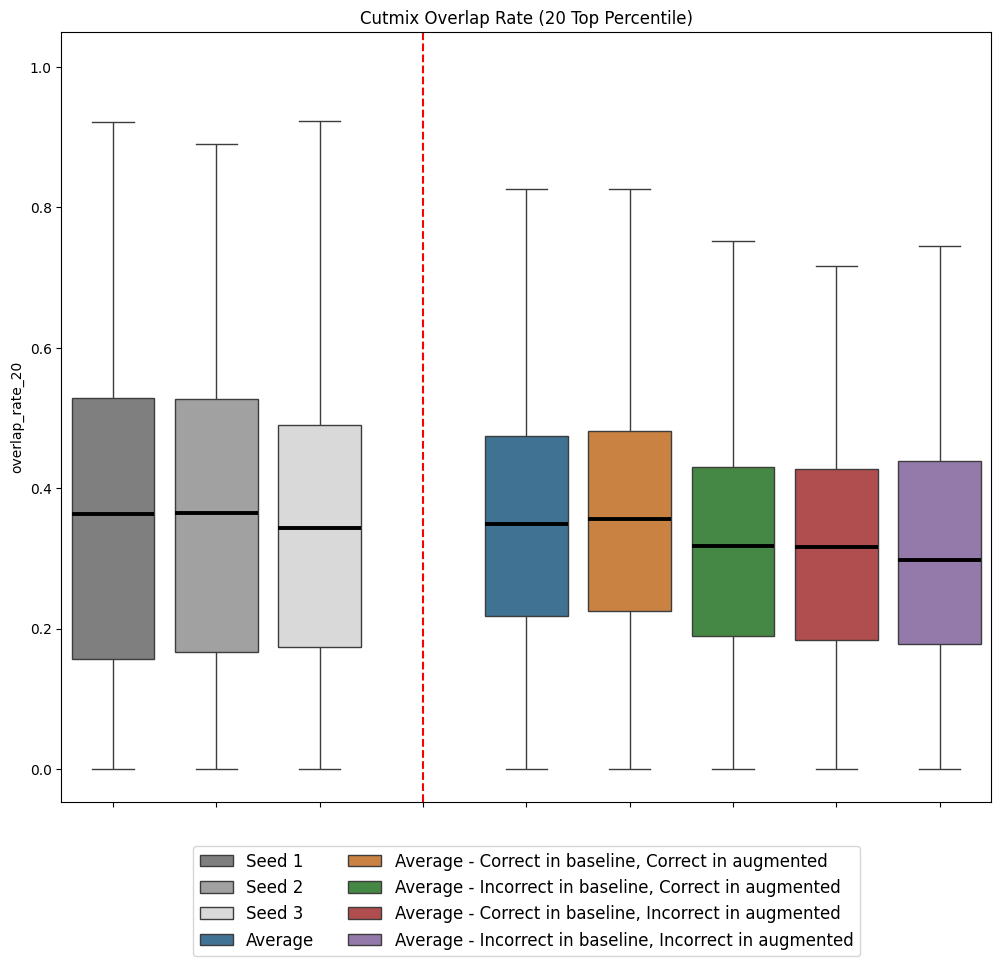}
\centering
\end{figure}

\begin{figure}[H]
\caption{Segmented boxplot for Elastic Transform and Overlap Rate (20), showing the distribution of the metric values across segmentations.}
\label{fig:seg:elastic}
\includegraphics[width=.75\linewidth]{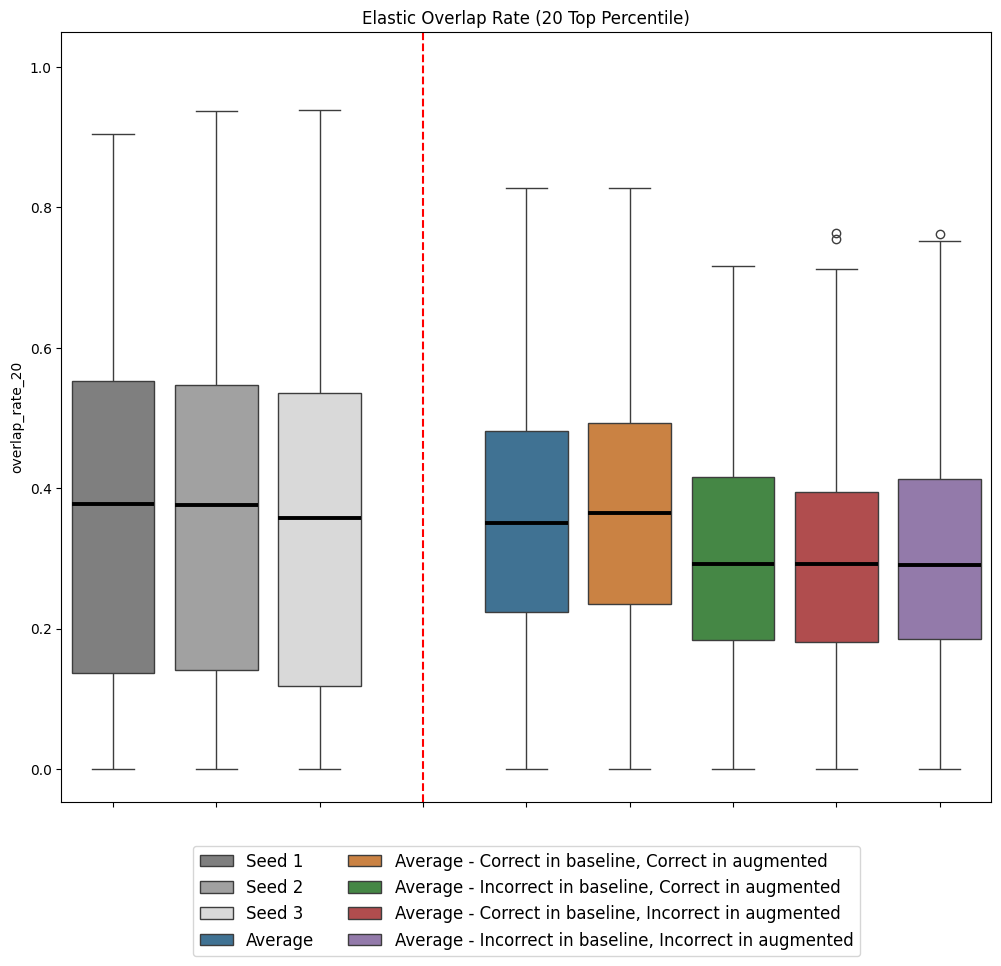}
\centering
\end{figure}

\begin{figure}[H]
\caption{Segmented boxplot for Equalization and Overlap Rate (20), showing the distribution of the metric values across segmentations.}
\label{fig:seg:eq}
\includegraphics[width=.75\linewidth]{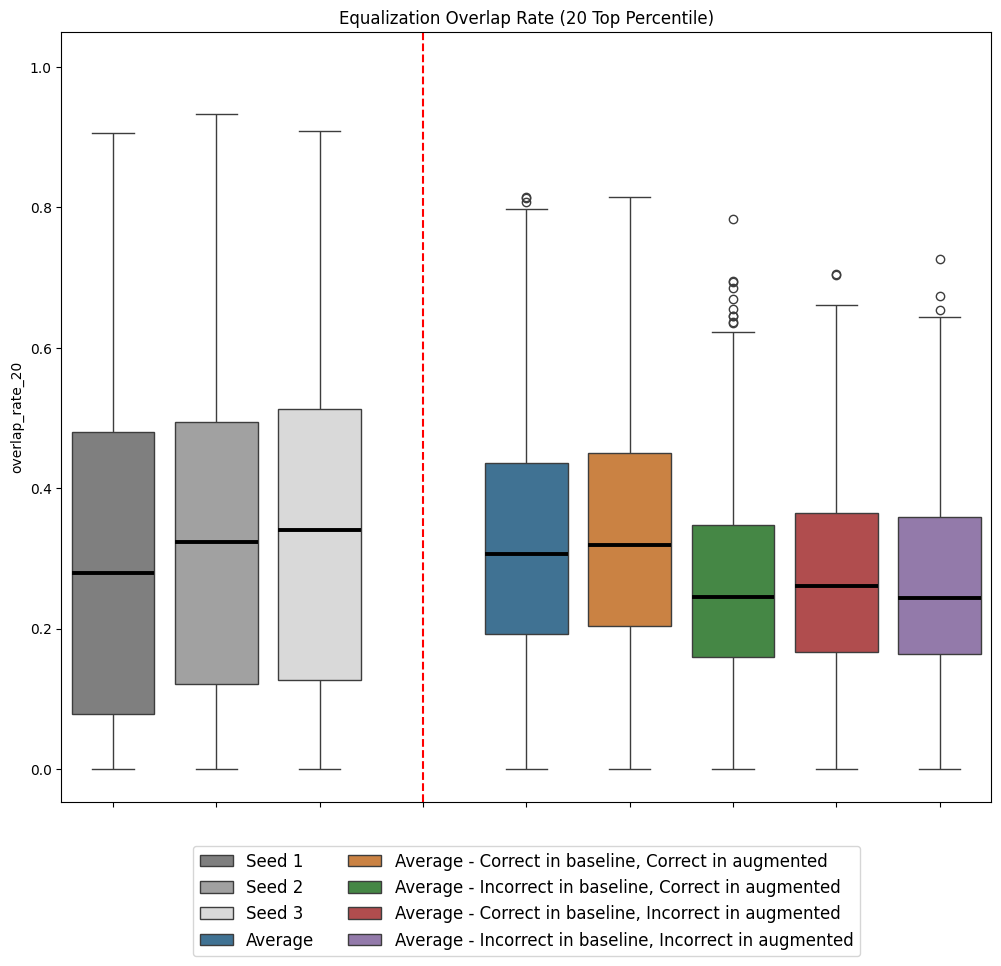}
\centering
\end{figure}

\begin{figure}[H]
\caption{Segmented boxplot for Gaussian Blur and Overlap Rate (20), showing the distribution of the metric values across segmentations.}
\label{fig:seg:gauss}
\includegraphics[width=.75\linewidth]{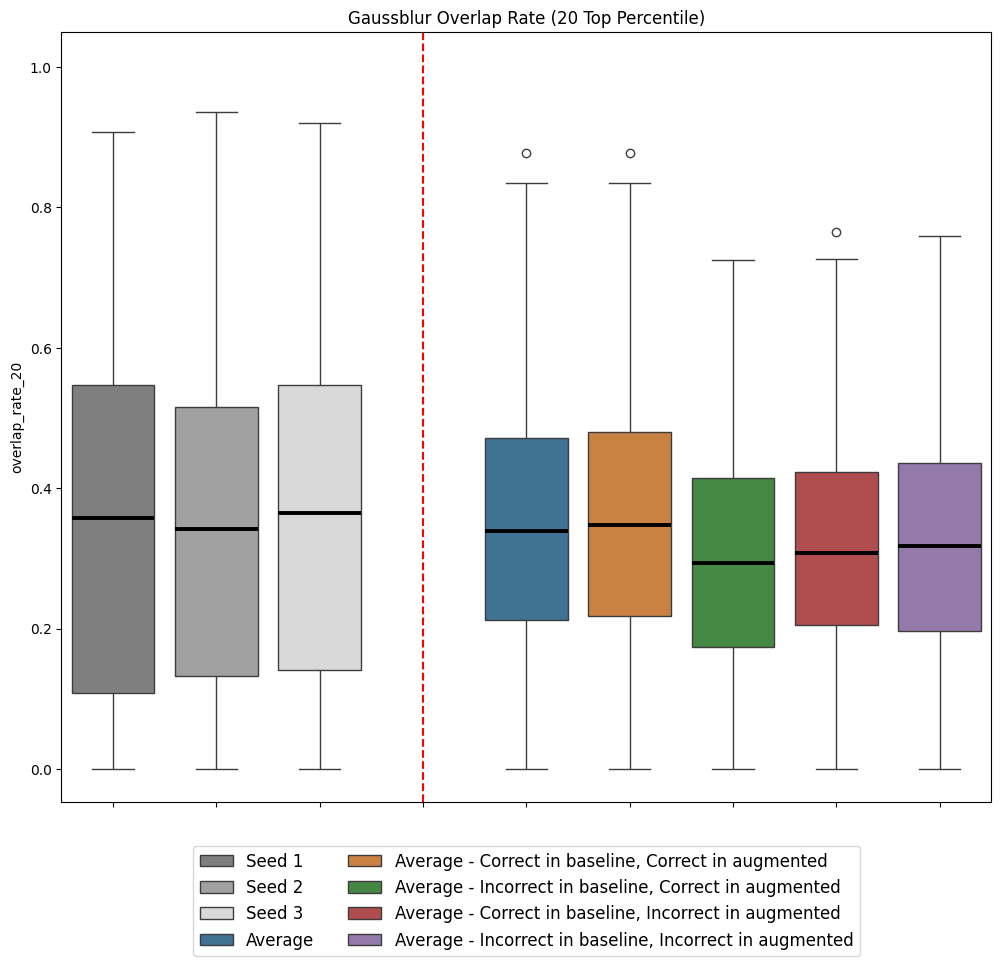}
\centering
\end{figure}

\begin{figure}[H]
\caption{Segmented boxplot for Random Cropping and Overlap Rate (20), showing the distribution of the metric values across segmentations.}
\label{fig:seg:rand}
\includegraphics[width=.75\linewidth]{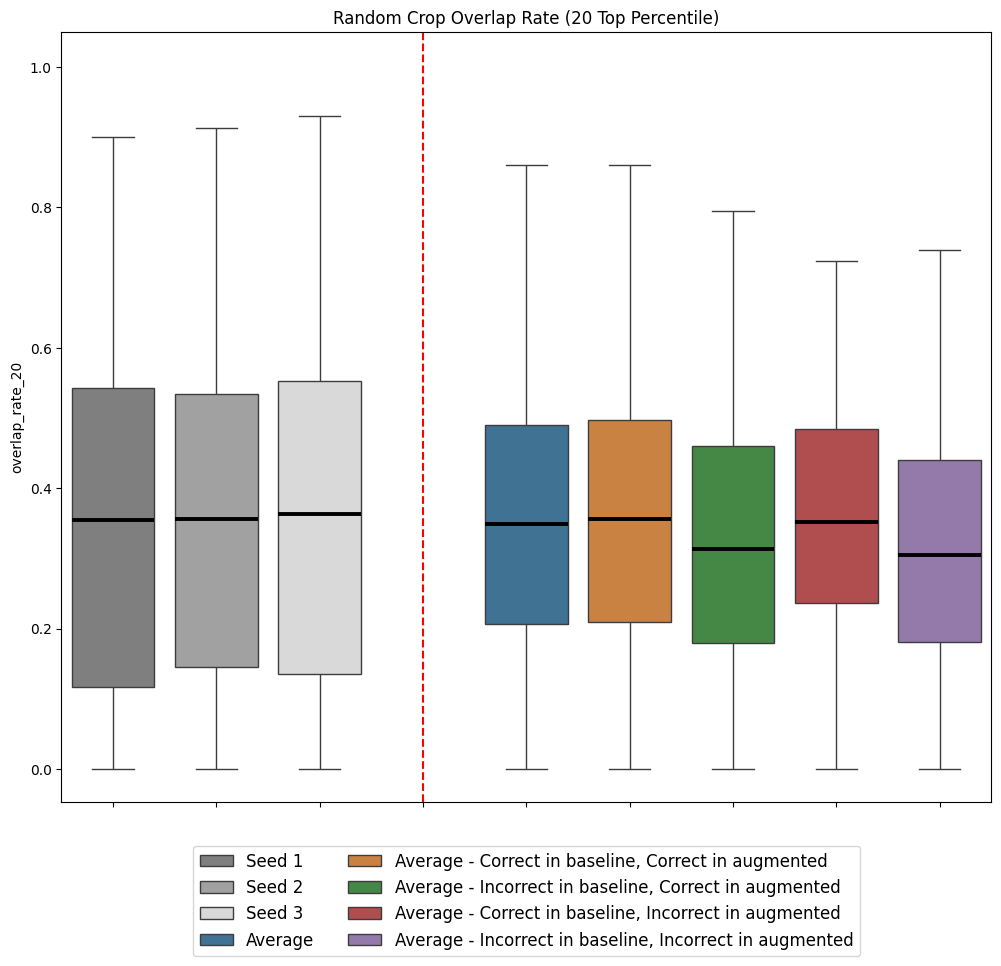}
\centering
\end{figure}

Figure \autoref{fig:seg:affine} shows the segmented boxplot for affine transform method, \autoref{fig:seg:colorjitter} shows it for color jitter, \autoref{fig:seg:cutmix} for cutmix, \autoref{fig:seg:elastic} for elastic transform, \autoref{fig:seg:eq} for equalization, \autoref{fig:seg:gauss} for Gaussian blur, and \autoref{fig:seg:rand} for random cropping

\end{document}